\documentclass[sort,compress,11pt]{elsarticle}

\usepackage{amssymb}
\usepackage{amsthm}
\usepackage{amsmath, esint}
\usepackage{mathtools}
\usepackage{mathrsfs}
\usepackage[ruled, lined, linesnumbered, longend]{algorithm2e}
\usepackage{array}
\usepackage{booktabs}
\usepackage{multirow}
\usepackage{listings}
\usepackage{tabu}
\usepackage{enumerate}
\usepackage{lineno}
\usepackage{fullpage}
\usepackage{xcolor}
\usepackage[colorinlistoftodos]{todonotes}
\usepackage[colorlinks=true]{hyperref}
\usepackage{url}
\usepackage{textcomp}
\usepackage{float}
\usepackage{graphicx}
\usepackage{subfig}

\usepackage{makecell}
\usepackage[export]{adjustbox}
\usepackage{caption}
\usetikzlibrary{shapes}
\biboptions{numbers,comma,round,square}
\usepackage[para,online,flushleft]{threeparttable}

\usepackage{orcidlink}

\definecolor{myblue}{RGB}{63, 144, 218}
\definecolor{myorange}{RGB}{221, 132, 82}
\definecolor{mygreen}{RGB}{85, 168, 104}
\definecolor{myyellow}{RGB}{255, 169, 14}
\definecolor{myred}{RGB}{189, 31, 1}

\graphicspath{ {./figs/} }

\journal{Elsevier}
\DontPrintSemicolon
\definecolor{orcidlogocol}{HTML}{A6CE39}
\begin{document}
\makeatletter
\def\ps@pprintTitle{%
  \let\@oddhead\@empty
  \let\@evenhead\@empty
  \let\@oddfoot\@empty
  \let\@evenfoot\@oddfoot
}
\makeatother
\begin{frontmatter}

\title{D-Flow SGLD: Source-Space Posterior Sampling for Scientific Inverse Problems with Flow Matching}

\author[cMAE]{Meet Hemant Parikh\orcidlink{0009-0008-1284-6368}}
\author[ndAME]{Yaqin Chen}
\author[cMAE,ndAME]{Jian-Xun Wang\orcidlink{0000-0002-9030-1733}}
\address[cMAE]{Sibley School of Mechanical and Aerospace Engineering, Cornell University, Ithaca, NY, USA}
\address[ndAME]{Department of Aerospace and Mechanical Engineering, University of Notre Dame, Notre Dame, IN, USA}


\begin{abstract}
Data assimilation and scientific inverse problems require reconstructing high-dimensional physical states from sparse and noisy observations, ideally with uncertainty-aware posterior samples that remain faithful to learned priors and governing physics. While training-free conditional generation is well developed for diffusion models, corresponding conditioning and posterior sampling strategies for Flow Matching (FM) priors remain comparatively under-explored, especially on scientific benchmarks where fidelity must be assessed beyond measurement misfit. In this work, we study training-free conditional generation for scientific inverse problems under FM priors and organize existing inference-time strategies by where measurement information is injected: (i) guided transport dynamics that perturb sampling trajectories using likelihood information, and (ii) source-distribution inference that performs posterior inference over the source variable while keeping the learned transport fixed. Building on the latter, we propose D-Flow SGLD, a source-space posterior sampling method that augments differentiable source inference with preconditioned stochastic gradient Langevin dynamics, enabling scalable exploration of the source posterior induced by new measurement operators without retraining the prior or modifying the learned FM dynamics. We benchmark representative methods from both families on a hierarchy of problems: 2D toy posteriors, chaotic Kuramoto–Sivashinsky trajectories, and wall-bounded turbulence reconstruction. Across these settings, we quantify trade-offs among measurement assimilation, posterior diversity, and physics/statistics fidelity, and establish D-Flow SGLD as a practical FM-compatible posterior sampler for scientific inverse problems.
\end{abstract}

\end{frontmatter}

\section{Introduction}

Scientific inference tasks such as data assimilation and inverse problems seek to reconstruct an unknown high-dimensional physical state from indirect, sparse, and noisy observations. In many settings, the forward map is known (e.g., a measurement operator, a numerical solver, or a sensor model), but the posterior distribution over states remains highly non-Gaussian and multimodal, making uncertainty quantification (UQ) computationally challenging for classical sampling-based Bayesian workflows \cite{biegler2010large}. Recently, deep generative models have emerged as powerful \emph{data-driven priors} that can represent complex distributions over scientific fields, trajectories, and turbulent fluctuations~\cite{du2024conditional,drygala2022generative,Parikh_Fan_Wang_2026}. Beyond unconditional generation, practical scientific use often requires \emph{conditional} generation: sampling states that are simultaneously (i) consistent with observations or constraints and (ii) plausible under the learned prior and, when applicable, the governing physics~\cite{chung2022diffusion,dasgupta2026unifying,du2024conditional}.

Score-based diffusion models and related denoising-based generators have enabled compelling results across scientific surrogates and conditional reconstruction \cite{shu2023physics,jacobsen2025cocogen,shan2025red,dong2025data,liu2025confild,gao2024bayesian}. A key appeal in inverse problems is the ability to reuse an unconditional prior and impose new observation models at test time, thereby avoiding retraining for each forward operator or sensing configuration \cite{chung2022diffusion,mardani2023variational}. This \emph{training-free} (or \emph{zero-shot}) conditioning paradigm is particularly attractive in scientific applications where paired training data for every measurement setup is unavailable and where one wishes to probe posterior uncertainty under changing experimental conditions~\cite{rozet2024learning,fan2025neural,du2025hug}.

Among recent generative paradigms, \emph{Flow Matching} (FM) offers a simulation-free training objective for continuous-time generative models by regressing a time-dependent vector field along a prescribed probability path between a simple source distribution and the data distribution \cite{lipman2022flow}. FM subsumes diffusion-type probability paths as special cases and also supports alternative path families (e.g., optimal transport-inspired constructions), which can lead to favorable practical trade-offs in training robustness and sampling efficiency depending on the chosen path and numerical solver~\cite{tong2023improving}. Despite this promise, FM remains less explored in scientific applications, with only a small but growing set of works on PDE-driven systems, physics-constrained sampling, and turbulence reconstruction~\cite{li2025generative,baldan2025flow,utkarsh2025physics,wang2025geofunflow,Parikh_Fan_Wang_2026}. More importantly for inverse problems, \emph{training-free} conditional generation, i.e., conditioning a fixed pretrained prior on new measurement operators without retraining, is well established for diffusion models~\cite{chung2022diffusion} but comparatively under-developed for FM~\cite{Parikh_Fan_Wang_2026}, especially under scientific fidelity requirements that go beyond measurement misfit. As a result, it is still unclear how existing FM-compatible conditioning strategies behave across operators and noise regimes, and which approaches best preserve physics-relevant structure and statistics. This gap motivates (i) a systematic benchmark of training-free conditioning methods within the FM framework and (ii) scalable FM-compatible posterior samplers tailored to scientific datasets.

Existing training-free approaches can be organized by \emph{where} measurement information is injected at inference time: (i) \emph{guided transport dynamics} (trajectory-level conditioning) versus (ii) \emph{source-distribution inference} (source-space conditioning). Both families aim to sample from a posterior induced by the observation model; they differ in whether the conditioning modifies the \emph{transport dynamics} during generation or instead performs inference over the \emph{source variable/distribution} while keeping the learned transport fixed. A first family conditions generation by perturbing the sampling dynamics using likelihood information, closely related to diffusion posterior sampling and guidance-based solvers~\cite{chung2022diffusion,mardani2023variational}. In practice, these methods modify the drift of the generative ODE/SDE (or its discretized update) so that the evolving sample is continuously pulled toward measurement-consistent regions, yielding an integration-with-guidance procedure rather than retraining the prior~\cite{pokle2023training,kim2025flowdps,zhang2025improving}. Representative instantiations include plug-and-play style flow guidance~\cite{martin2024pnp} and annealing-based schedules that control the strength of likelihood injection along the trajectory~\cite{zhang2025improving}. Recent extensions also incorporate gradient-free or particle-based mechanisms to improve robustness and diversity~\cite{kelvinius2025solving,rozet2024learning}. While effective in many imaging restoration problems, dynamics-level guidance can be delicate in scientific settings: aggressive enforcement of measurement consistency may drive intermediate states into regions poorly supported by the learned prior, and can complicate the preservation of physically meaningful structure across scales. A second family keeps the learned transport dynamics fixed and instead performs inference over the source field that initializes the transport map, seeking a source posterior whose pushforward yields measurement-consistent states~\cite{wang2024dmplug,ben2024d,purohit2025consistency,wang2025source,kalaivanan2025ess}. Approaches such as D-Flow optimize the source by differentiating through the ODE solver~\cite{ben2024d}, while related methods leverage consistency models for efficient source optimization~\cite{wang2024dmplug}. More recent works have explored source guidance using stochastic dynamics~\cite{purohit2025consistency} or gradient-free inference~\cite{kalaivanan2025ess} to estimate the source posterior. Theoretically, this paradigm is attractive for scientific problems because conditional samples remain exact outputs of the pretrained transport, but it introduces practical trade-offs between the cost of repeated forward/adjoint solves and the need for well-mixed, diverse posterior ensembles.

In this work, we study training-free conditional generation for scientific inverse problems in the FM framework and address the lack of scalable posterior sampling methods that preserve physics-relevant structure. Concretely, we (i) provide a systematic benchmark of representative training-free conditioning strategies for FM priors, covering both guided transport dynamics and source-distribution inference, and evaluate them under complementary criteria that distinguish measurement assimilation from prior/physics fidelity; and (ii) introduce \emph{D-Flow SGLD}, a source-space posterior sampling approach that augments differentiable source inference with \emph{preconditioned} stochastic gradient Langevin dynamics (SGLD)~\cite{li2016preconditioned}, enabling efficient exploration of the source posterior induced by new measurement operators without modifying the learned FM dynamics or retraining the prior. We validate these ideas across a hierarchy of inverse problems, including 2D toy problems with a high-quality reference posterior, chaotic Kuramoto–Sivashinsky trajectories with PDE-residual fidelity, and wall-bounded turbulence reconstruction with spectral diagnostics, thereby clarifying when dynamics-level guidance versus source-space inference is more reliable and providing a practical FM-compatible posterior sampler for scientific inverse problems.

\section{Methodology}

We study \emph{training-free} Bayesian inverse problems using a pretrained, unconditional flow matching prior. Given a generative transport map learned from representative data, our goal is to approximate the posterior distribution induced by partial and noisy observations without retraining the generative model. We organize inference-time conditioning according to \emph{where} measurement information enters the generative procedure: (i) \emph{velocity-field guidance}, which modifies the transport dynamics during sampling using likelihood-driven corrections; and (ii) \emph{source-space guidance}, which keeps the learned transport unchanged and instead performs inference over the source variables, pushing inferred sources forward through the fixed map. Within the source-space family, existing approaches are typically deterministic (MAP) optimizers, such as D-Flow. We introduce a stochastic variant, \emph{D-Flow SGLD}, which targets the source posterior via SGLD and returns posterior samples by a single pushforward. 

\subsection{Bayesian inverse problems with a flow-matching prior}
\label{subsec:general_inverse}

Given a physical state $\mathbf{x}_1\in\mathbb{R}^n$ (e.g., 2D/3D velocity fields) and noisy, partial measurements $\mathbf{y}\in\mathbb{R}^m$ with $m \ll n$, we seek to reconstruct $\mathbf{x}_1$ in a manner that is both physically plausible and consistent with $\mathbf{y}$.
We assume a (possibly nonlinear) state-to-observable operator $\mathcal{F}:\mathbb{R}^n \to \mathbb{R}^m$ with additive noise,
\begin{equation}
\label{eq:meas_general}
\mathbf{y}=\mathcal{F}(\mathbf{x}_1)+\boldsymbol{\epsilon}, 
\qquad \boldsymbol{\epsilon}\sim p_{\boldsymbol{\epsilon}},
\end{equation}
which defines a likelihood $p(\mathbf{y}|\mathbf{x}_1)$. Common instances include masked or cropped observations, sparse sensors, linear projections, and nonlinear derived quantities (e.g., vorticity). For Gaussian noise $\boldsymbol{\epsilon} \sim \mathcal{N}(0, \sigma_y^2I)$, $-\log p(\mathbf{y}|\mathbf{x}_1) = (2\sigma^2_y)^{-1}\|\mathbf{y} - \mathcal{F}(\mathbf{x}_1)\|^2_2 + \mathrm{const}$. The Bayesian inverse problem is to characterize the posterior,
\begin{equation}
\label{eq:posterior_general}
p(\mathbf{x}_1|\mathbf{y}) \propto p(\mathbf{y}|\mathbf{x}_1)p(\mathbf{x}_1).
\end{equation}
where $p_1 = p(\mathbf{x}_1)$ is a learned prior over physically plausible states. In this work we consider both (i) point estimation via the maximum a posteriori (MAP) solution,
\begin{equation}
 \mathbf{x}_{\mathrm{MAP}}\in\arg\min_{\mathbf{x}_1}\big(-\log p(\mathbf{y}|\mathbf{x}_1)-\log p(\mathbf{x}_1)\big),   
\end{equation}
and (ii) posterior sampling for uncertainty quantification and downstream statistics.

\subsection{Flow matching prior: transport map learned by (OT-)CFM}
\label{subsec:flow_matching_prior}

We parameterize the learned prior $p_1$ via a deterministic transport map
$\mathcal{T}_\theta$ defined by the ODE
\begin{equation}
\label{eq:transport_general}
\frac{d\mathbf{x}_\tau}{d\tau}=\boldsymbol{\nu}_\theta(\tau,\mathbf{x}_\tau),
\qquad 
\mathbf{x}_0\sim p_0,\ \mathbf{x}_1=\mathcal{T}_\theta(\mathbf{x}_0),\ \tau\in[0,1],
\end{equation}
so that $p_1 = p(\mathbf{x}_1)$ is a prior approximation of data distribution. In all inverse problems considered below, $\mathcal{T}_\theta$ (equivalently $\boldsymbol{\nu}_\theta$) is pretrained once on representative unconditional data and then held fixed. We train
$\boldsymbol{\nu}_\theta$ using CFM \cite{lipman2022flow} and its mini-batch optimal-transport variant (OT-CFM)~\cite{tong2023improving}.

\paragraph{Flow matching and conditionalization}
Let $\{p_\tau\}_{\tau\in[0,1]}$ denote a (generally intractable) marginal path of densities connecting $p_0$ and $p_1$, with associated marginal velocity field $\boldsymbol{\nu}(\tau,\mathbf{x})$ satisfying the continuity equation. Flow matching learns $\boldsymbol{\nu}_\theta$ by matching this marginal velocity in mean square,
\begin{equation}
\label{eq:fm_obj}
\mathcal{L}_{\mathrm{FM}}(\theta) =
\mathbb{E}_{\tau\sim\mathcal{U}[0,1],\ \mathbf{x}_\tau\sim p_\tau}
\Big[\big\|\boldsymbol{\nu}_\theta(\tau,\mathbf{x}_\tau)-\boldsymbol{\nu}(\tau,\mathbf{x}_\tau)\big\|_2^2\Big].
\end{equation}
The difficulty is that both sampling $\mathbf{x}_\tau\sim p_\tau$ and evaluating the marginal velocity $\boldsymbol{\nu}$ are intractable. CFM addresses this by representing the marginal path as a \emph{mixture} of tractable conditional paths. Namely, an auxiliary variable $\mathbf{z} \sim q(\mathbf{z})$ and conditional densities $p_\tau(\cdot|\mathbf{z})$ are introduced such that,
\begin{equation}
\label{eq:mixture_path}
p_\tau(\mathbf{x}) = \int p_\tau(\mathbf{x}|\mathbf{z})q(\mathbf{z})d\mathbf{z}.
\end{equation}
Each conditional path is equipped with a known conditional velocity $\boldsymbol{\nu}(\tau,\mathbf{x}|\mathbf{z})$ (chosen by design). For mixtures of continuity-equation solutions, the induced marginal velocity satisfies the standard identity~\cite{lipman2022flow}
\begin{equation}
\label{eq:mixture_velocity}
\boldsymbol{\nu}(\tau,\mathbf{x}) = \mathbb{E}_{\mathbf{z}\sim q(\mathbf{z} | \mathbf{x},\tau)}\big[\boldsymbol{\nu}(\tau,\mathbf{x} | \mathbf{z})\big],
\qquad
q(\mathbf{z} | \mathbf{x},\tau)=\frac{p_\tau(\mathbf{x} | \mathbf{z})q(\mathbf{z})}{p_\tau(\mathbf{x})},
\end{equation}
i.e., the marginal velocity is the posterior expectation of the conditional velocity under the mixing variable. CFM then replaces the intractable regression target $\boldsymbol{\nu}(\tau,\mathbf{x})$ in \eqref{eq:fm_obj} with the tractable conditional target $\boldsymbol{\nu}(\tau,\mathbf{x} | \mathbf{z})$ by sampling from the conditional paths:
\begin{equation}
\label{eq:cfm_obj}
\mathcal{L}_{\mathrm{CFM}}(\theta)
=
\mathbb{E}_{\tau\sim\mathcal{U}[0,1],\ \mathbf{z} \sim q(\mathbf{z}),\ \mathbf{x}_\tau\sim p_\tau(\cdot\mid\mathbf{z})}
\Big[\big\|\boldsymbol{\nu}_\theta(\tau,\mathbf{x}_\tau)-\boldsymbol{\nu}(\tau,\mathbf{x}_\tau|\mathbf{z})\big\|_2^2\Big].
\end{equation}
Under the regularity conditions in \cite{lipman2022flow}, $\nabla_\theta\mathcal{L}_{\mathrm{CFM}}$ matches
$\nabla_\theta\mathcal{L}_{\mathrm{FM}}$ up to a constant, so minimizing \eqref{eq:cfm_obj} yields a consistent estimator of the marginal velocity field along $\{p_\tau\}$ while avoiding likelihood evaluation and Jacobian-trace computation.

\paragraph{OT-CFM instantiation}
CFM admits substantial flexibility in the choice of the auxiliary variable $\mathbf{z}\sim q(\mathbf{z})$ and the associated conditional path family $\{p_\tau(\cdot|\mathbf{z}),\boldsymbol{\nu}(\tau,\cdot|\mathbf{z})\}$. In general, $z$ can encode (i) independently sampled endpoint pairs $(\mathbf{x}_0,\mathbf{x}_1)$, (ii) structured couplings between $p_0$ and $p_1$ (e.g., optimal-transport plans), or (iii) other latent constructions that yield tractable intermediate distributions and known conditional velocities~\cite{tong2023improving}. These choices affect the geometry of the learned transport; in particular, stronger endpoint
couplings tend to bias training toward straighter trajectories and faster ODE sampling.

In this work we use the mini-batch 2-Wasserstein OT coupling of Tong et al.~\cite{tong2023improving}. Concretely, we set the auxiliary variable to the pair of endpoints $\mathbf{z} = (\mathbf{x}_0,\mathbf{x}_1)$ sampled from a mini-batch OT plan
$\pi(\mathbf{x}_0,\mathbf{x}_1)$. For each mini-batch we form an approximate 2-Wasserstein coupling between source samples $\{\mathbf{x}_0^{(i)}\}_{i=1}^B\sim p_0$ and data samples $\{\mathbf{x}_1^{(i)}\}_{i=1}^B\sim p_1$, sample
$(\mathbf{x}_0,\mathbf{x}_1)\sim \pi$, and define a Gaussian bridge with linearly interpolated mean,
\begin{subequations}
\label{eq:otcfm_bridge}
\begin{align}
p_\tau(\mathbf{x}_\tau | \mathbf{x}_0,\mathbf{x}_1)
&=
\mathcal{N}\big((1-\tau)\mathbf{x}_0+\tau\mathbf{x}_1,\ \sigma^2\mathbf{I}\big), \\
\boldsymbol{\nu}(\tau,\mathbf{x}_\tau | \mathbf{x}_0,\mathbf{x}_1)
&=
\mathbf{x}_1-\mathbf{x}_0.
\end{align}
\end{subequations}
We optimize \eqref{eq:cfm_obj} by sampling $\mathbf{x}_\tau\sim p_\tau(\cdot\mid \mathbf{x}_0,\mathbf{x}_1)$. The OT
pairing biases training toward low-cost couplings and empirically yields near-straight transport trajectories, which
reduces the numerical stiffness of \eqref{eq:transport_general} and decreases the number of ODE function evaluations
required to sample from the pretrained prior. Training details (mini-batch OT solver, $\sigma$, network architecture,
and optimizer settings) are reported in Section~\ref{sec:arch-and-hyperparams}.

\subsection{Training-free conditioning I: guided transport dynamics}
\label{subsec:vel_guidance}

Velocity-field guidance targets training-free conditional sampling from the posterior $p(\mathbf{x}_1| \mathbf{y})$ by modifying the sampling dynamics of a fixed, pretrained OT-CFM prior at inference time. The guiding principle parallels posterior score guidance in diffusion models: when a sampler can be written as a probability-flow ODE, replacing the prior score
by the posterior score $\nabla_{\mathbf{x}_1}\log p(\mathbf{x}_1|\mathbf{y})
=\nabla_{\mathbf{x}_1}\log p(\mathbf{x}_1)+\nabla_{\mathbf{x}_1}\log p(\mathbf{y}|\mathbf{x}_1)$ introduces an additive likelihood-driven correction \cite{song2020score,chung2022diffusion,chung2025diffusion}. We adopt the same structure in
the flow-matching setting by augmenting the OT-CFM transport velocity with an approximate likelihood-gradient term. A derivation of the score/flow connection and its relation to probability-flow dynamics is provided in~\ref{app:score_flow_connection}.

\paragraph{Guided transport ODE.}
Let $\boldsymbol{\nu}_\theta(\tau,\mathbf{x})$ denote the pretrained OT-CFM velocity field defining the unconditional prior transport \eqref{eq:transport_general}. We define the guided dynamics
\begin{equation}
\label{eq:guided_ode_main}
\frac{d\mathbf{x}_\tau}{d\tau}
=
\boldsymbol{\nu}_\theta(\tau,\mathbf{x}_\tau)
+
\lambda(\tau)\mathbf{g}(\tau,\mathbf{x}_\tau;\mathbf{y}),
\qquad \tau\in[0,1],
\end{equation}
where $\lambda(\tau)\ge 0$ is a guidance schedule and $\mathbf{g}$ is chosen to increase measurement consistency (likelihood) while remaining compatible with the learned prior transport. 
The correction field $\mathbf{g}$ is not unique in general. We choose it to align with the likelihood score, since $\nabla_{\mathbf{x}_1}\log p(\mathbf{y}|\mathbf{x}_1)$ gives the steepest-ascent direction for increasing data likelihood, and it arises naturally as the additional term when inserting the posterior-score decomposition into probability-flow
ODEs for diffusion models (see derivation in \ref{app:score_flow_connection}). In practice, because $\mathcal{F}$ is defined on the clean state while $\mathbf{x}_\tau$ is an intermediate transport state, we evaluate the likelihood gradient at a denoised predictor $\widehat{\mathbf{x}}_{1|\tau}$ constructed from $\mathbf{x}_\tau$.
Assuming Gaussian measurement noise, $p(\mathbf{y}|\mathbf{x}_1)=\mathcal{N}(\mathcal{F}(\mathbf{x}_1),\sigma_y^2\mathbf{I})$, we have
\begin{equation}
\label{eq:lik_score_main}
\nabla_{\mathbf{x}_1}\log p(\mathbf{y}|\mathbf{x}_1) =
-\frac{1}{\sigma_y^2}\nabla_{\mathbf{x}_1}\|\mathbf{y}-\mathcal{F}(\mathbf{x}_1)\|_2^2.
\end{equation}
For OT-CFM, we use a straight-path one-step predictor
\begin{equation}
\label{eq:predictor_main}
\widehat{\mathbf{x}}_{1\mid\tau} = \mathbf{x}_\tau + (1-\tau)\boldsymbol{\nu}_\theta(\tau,\mathbf{x}_\tau),
\end{equation}
and define the correction direction by the plug-in gradient,
\begin{equation}
\label{eq:proxy_score_main}
\mathbf{g}(\tau,\mathbf{x}_\tau;\mathbf{y})
\ \propto\
\nabla_{\mathbf{x}_\tau}\log p(\mathbf{y}\mid \widehat{\mathbf{x}}_{1\mid\tau})
=
-\frac{1}{\sigma_y^2}\nabla_{\mathbf{x}_\tau}\left\|\mathbf{y}-\mathcal{F}\left(\widehat{\mathbf{x}}_{1\mid\tau}\right)\right\|_2^2.
\end{equation}
Gradients are obtained by autodiff through $\mathcal{F}$ and (optionally) through the predictor~\eqref{eq:predictor_main}.

\paragraph{Grad-free vs.\ Grad implementations}
Equation \eqref{eq:proxy_score_main} may backpropagate through $\nu_\theta$ inside $\widehat{\mathbf{x}}_{1\mid\tau}$.
We report two implementations:
(i) \emph{Grad-free}, which stops gradients through $\nu_\theta$ in \eqref{eq:predictor_main} (lower cost, biased
correction), and
(ii) \emph{Grad}, which differentiates through $\nu_\theta$ (higher cost, typically sharper conditioning).

\paragraph{Normalized guidance and tuning}
To stabilize tuning across inverse problems, we normalize the data-misfit gradient direction and scale it by a user-defined strength $b$ and the magnitude of the unconditional velocity:
\begin{equation}
\label{eq:guidance_main}
\mathbf{g}(\tau,\mathbf{x}_\tau;\mathbf{y})
= - b\|\boldsymbol{\nu}_\theta(\tau,\mathbf{x}_\tau)\|_2\;
\frac{\nabla_{\mathbf{x}_\tau}\left\|\mathbf{y}-\mathcal{F}\left(\widehat{\mathbf{x}}_{1\mid\tau}\right)\right\|_2^2}
{\left\|\nabla_{\mathbf{x}_\tau}\left\|\mathbf{y}-\mathcal{F}\left(\widehat{\mathbf{x}}_{1\mid\tau}\right)\right\|_2^2\right\|_2+\varepsilon},
\end{equation}
with $\varepsilon>0$ preventing division by zero. The $\lambda(\tau)$ and $b$ are reported in Section~\ref{sec:arch-and-hyperparams}.


\subsection{Training-free conditioning II: source-space posterior sampling (D-Flow SGLD)}

Velocity-field guidance (Section~\ref{subsec:vel_guidance}) enforces measurement consistency by perturbing the
pretrained transport dynamics. While effective and fully training-free, this strategy introduces practical trade-offs: conditional trajectories depend on the guidance schedule and strength, and the correction direction is typically constructed from plug-in likelihood gradients evaluated at a denoised predictor. As a result, aggressive guidance can bias the sampling path, increase path length (and hence numerical stiffness), and steer intermediate states toward regions that are less supported by the learned prior. These considerations motivate a complementary conditioning mechanism that \emph{keeps the pretrained transport map fixed} and incorporates measurements only through inference over the source variables.

\paragraph{Source-space posterior induced by a fixed OT-CFM prior}
Recall the pretrained transport representation \eqref{eq:transport_general}:
$\mathbf{x}_0\sim p_0$ and $\mathbf{x}_1=\mathcal{T}_\theta(\mathbf{x}_0)$, with $p_1=(\mathcal{T}_\theta)_\#p_0$.
Conditioning on observations $\mathbf{y}$ with likelihood $p(\mathbf{y}\mid \mathbf{x}_1)$ induces a posterior over the
source variable,
\begin{equation}
\label{eq:source_posterior}
p(\mathbf{x}_0|\mathbf{y}) \ \propto\ p\left(\mathbf{y}|\mathcal{T}_\theta(\mathbf{x}_0)\right)p_0(\mathbf{x}_0).
\end{equation}
This \emph{source-space guidance} preserves the learned transport geometry and generates conditional samples by pushforward, 
\begin{equation}
\label{eq:source_pushforward}
\mathbf{x}_1^{(s)}=\mathcal{T}_\theta(\mathbf{x}_0^{(s)}),
\qquad
\mathbf{x}_0^{(s)}\sim p(\mathbf{x}_0 | \mathbf{y}).
\end{equation}

\paragraph{Deterministic source inference: D-Flow}
A deterministic instance of \eqref{eq:source_posterior} is obtained by computing the MAP estimate in source space,
\begin{equation}
\label{eq:dflow_map}
\mathbf{x}_0^\star
\in
\arg\min_{\mathbf{x}_0}
\Big[
-\log p\left(\mathbf{y}| \mathcal{T}_\theta(\mathbf{x}_0)\right)
-\log p_0(\mathbf{x}_0)
\Big],
\qquad
\mathbf{x}_1^\star=\mathcal{T}_\theta(\mathbf{x}_0^\star),
\end{equation}
which corresponds to D-Flow~\cite{ben2024d}. The objective in \eqref{eq:dflow_map} can be viewed as a data-misfit term
(e.g., induced by $\mathcal{F}$ and the noise model) plus an explicit regularization in the source space given by the
negative log source prior. For the common choice $p_0=\mathcal{N}(\mathbf{0},\mathbf{I})$, we have
$-\log p_0(\mathbf{x}_0)=\tfrac{1}{2}\|\mathbf{x}_0\|_2^2+\mathrm{const}$, yielding an $\ell_2$ regularizer that keeps
the inferred source sample within typical-set regions of the pretrained generator, but is ignored in practice as the authors demonstrate that D-Flow is implicitly regularized by minimizing the negative log-likelihood as a function of the source sample $x_0$~\cite{ben2024d}. More generally, one may include
additional problem-dependent penalties (e.g., smoothness, sparsity, or physics-inspired constraints) either as an
augmented source regularizer or as a penalty on the reconstructed state:
\begin{equation}
\label{eq:dflow_map_reg}
\mathbf{x}_0^\star
\in
\arg\min_{\mathbf{x}_0}
\Big[
-\log p\left(\mathbf{y}|\mathcal{T}_\theta(\mathbf{x}_0)\right)
-\log p_0(\mathbf{x}_0)
+\alpha\mathcal{R}_0(\mathbf{x}_0)
+\beta\mathcal{R}_1\big(\mathcal{T}_\theta(\mathbf{x}_0)\big)
\Big],
\end{equation}
with weights $\alpha,\beta\ge 0$. In practice, \eqref{eq:dflow_map} (or \eqref{eq:dflow_map_reg} when used) is solved by
gradient-based optimization, requiring differentiation through the measurement operator $\mathcal{F}$ and the fixed
transport map $\mathcal{T}_\theta$ (i.e., through the ODE solve defined by $\boldsymbol{\nu}_\theta$). D-Flow returns a
single high-probability reconstruction (a posterior mode) and therefore does not provide posterior uncertainty.

\paragraph{D-Flow SGLD for posterior sampling}
To move beyond the point estimator \eqref{eq:dflow_map} and quantify posterior uncertainty, we propose \emph{D-Flow SGLD}, a training-free MCMC method that targets the full source posterior \eqref{eq:source_posterior} and produces state-space samples by a single pushforward through the fixed transport map $\mathcal{T}_\theta$.

We define a source-space energy whose Boltzmann distribution equals the desired posterior:
\begin{equation}
\label{eq:source_energy}
U(\mathbf{x}_0;\mathbf{y})
=
-\log p\left(\mathbf{y}|\mathcal{T}_\theta(\mathbf{x}_0)\right)
-\log p_0(\mathbf{x}_0)
+\alpha\mathcal{R}_0(\mathbf{x}_0)
+\beta\mathcal{R}_1\big(\mathcal{T}_\theta(\mathbf{x}_0)\big),
\end{equation}
so that $p(\mathbf{x}_0|\mathbf{y}) \propto \exp\{-U(\mathbf{x}_0;\mathbf{y})\}$. For $p_0=\mathcal{N}(\mathbf{0},\mathbf{I})$, the prior contributes $-\log p_0(\mathbf{x}_0)=\tfrac{1}{2}\|\mathbf{x}_0\|_2^2+\mathrm{const}$, which regularizes inference toward typical-set regions of the pretrained generator and stabilizes optimization in high-dimensional source spaces.

We sample from $\exp\{-U(\mathbf{x}_0;\mathbf{y})\}$ using preconditioned stochastic gradient Langevin dynamics (SGLD)~\cite{li2016preconditioned}:
\begin{equation}
\label{eq:dflow_sgld}
\mathbf{x}_0^{k+1}
=
\mathbf{x}_0^{k}
-\eta_k\,\mathbf{P}_k\nabla_{\mathbf{x}_0}U(\mathbf{x}_0^{k};\mathbf{y})
+\sqrt{2\eta_k}\mathbf{P}_k^{1/2}\boldsymbol{\xi}_k,
\qquad
\boldsymbol{\xi}_k\sim \mathcal{N}(\mathbf{0},\mathbf{I}),
\end{equation}
where $\{\eta_k\}$ is a stepsize schedule and $\mathbf{P}_k\succ 0$ is a (diagonal) adaptive preconditioner computed from running second-moment statistics of $\nabla_{\mathbf{x}_0}\log p(\mathbf{y}|\mathcal{T}_\theta(\mathbf{x}_0))$. The preconditioner is applied consistently to both the drift and diffusion terms to improve mixing when different source coordinates exhibit highly anisotropic curvature. Each evaluation of $\nabla_{\mathbf{x}_0}U$ differentiates through the measurement operator $\mathcal{F}$ and the fixed transport map $\mathcal{T}_\theta$ (i.e., through the ODE solve defined by $\boldsymbol{\nu}_\theta$), but no retraining of $\boldsymbol{\nu}_\theta$ is required.
In this work, $M$ i.i.d. samples from a standard Gaussian distribution, $\mathbf{x}_0^{(j),0} \sim \mathcal{N}(0, \mathbf{I})$, initialize $M$ parallel chains for D-Flow SGLD. To reduce burn-in and improve robustness under multi-modal posteriors, SGLD can be warm-started from partially optimized deterministic source MAP solutions. Concretely, we run a few steps of $M$ independent MAP optimizations with distinct random seeds to obtain a set of candidate modes, $\{\mathbf{x}_{0,\mathrm{MAP}}^{(j)}\}_{j=1}^M$. These modes initialize the $M$ SGLD chains as:
\begin{equation}
\label{eq:ensemble_init}
\mathbf{x}0^{(j),0} = \mathbf{x}{0,\mathrm{MAP}}^{(j)}, \qquad j=1,\dots,M.
\end{equation}
Although it incurs a slightly higher computational cost, this ``deep-ensemble'' approach yields parallel chains that explore distinct high-density basins of multi-modal posteriors while retaining a shared target energy across all chains.

After burn-in, we collect source samples $\{\mathbf{x}_0^{k}\}_{k\in\mathcal{K}}$ (optionally thinned to reduce autocorrelation) and return conditional samples in state space by pushforward:
\begin{equation}
\label{eq:dflow_sgld_pushforward}
\mathbf{x}_1^{(j),k}=\mathcal{T}_\theta(\mathbf{x}_0^{(j),k}),
\qquad j = 1, \cdots, M, \quad k\in\mathcal{K}.
\end{equation}
We refer to \eqref{eq:dflow_sgld}--\eqref{eq:dflow_sgld_pushforward} as \emph{D-Flow SGLD}. Pseudocode and hyperparameter settings (burn-in, thinning, stepsize schedule, and preconditioner construction) are provided in~\ref{sec:d-flow-sgld-algo} and~\ref{sec:arch-and-hyperparams} respectively.

\section{Numerical results}

\subsection{Data generation}
\label{sec:data_generation}

We evaluate training-free conditioning on three datasets spanning increasing dimensionality and physical complexity. In all cases, the generative prior is trained on an unconditional training set, while conditional generation are performed on a held-out validation set from the same distribution.

\paragraph{Toy distributions}
We consider two standard two-dimensional benchmarks: the \emph{S-curve} and the \emph{two-moons} distribution. For each dataset, we draw $10{,}240$ i.i.d.\ samples to train the flow-matching prior; conditional experiments use a held-out validation set. Unless otherwise noted, samples are perturbed by isotropic Gaussian noise $\epsilon\sim\mathcal{N}(0,\sigma^2\mathbf{I})$. Complete sampling details (parameterizations and scaling) are provided in~\ref{app:toy_data}.

\paragraph{Kuramoto--Sivashinsky (KS) dynamics}
We generate spatio-temporal solution trajectories $u(x,t)$ of the one-dimensional KS equation on a periodic domain of length $L=20\pi$ using the spectral solver in JAX-CFD~\cite{Dresdner2022-Spectral-ML}. Each trajectory is stored on a uniform grid of size $N_x\times N_t = 256\times 256$ with time step $\Delta t=0.2$. Initial conditions $u(x,0)$ are sampled as random linear combinations of sinusoidal modes with random phases and amplitudes. In total we generate $10{,}000$ trajectories and split them into training/validation sets for prior learning and conditional evaluation, respectively.

\paragraph{Turbulent channel flow (DNS)}
We use an incompressible channel-flow DNS at friction Reynolds number $Re_\tau = 180$ computed with our in-house GPU CFD solver Diff-FlowFSI~\cite{fan2025diff}. From the simulation we extract $21{,}900$ snapshots of the velocity fluctuations $\mathbf{u}'=[u',v',w']$ at wall-normal location $y^+=40$, sampled with non-dimensional interval $\Delta t^+=0.4$. Each snapshot is downsampled from the original $n_x\times n_z=320\times 200$ grid to $128\times 128$
using bicubic interpolation. Conditional generation is evaluated on held-out validation sets not used for training.

\subsection{Evaluation Metrics}
\label{sec:eval-metrics}
We evaluate conditional generation using two complementary criteria: \emph{(i) measurement assimilation}, i.e., whether generated samples are consistent with the observed data under the measurement operator; and \emph{(ii) prior/physics fidelity}, i.e., whether conditional samples remain plausible under the learned data distribution and (when available) the governing physics. The first criterion is common across all datasets, while the second is case-specific because ``fidelity'' has different operational definitions for toy distributions, PDE trajectories, and turbulent fields.

\paragraph{Measurement assimilation (data misfit)}
For each validation instance with observation $\mathbf{y}$, we generate $N$ conditional samples $\{\hat{\mathbf{x}}_1^{(i)}\}_{i=1}^N$ and compute predicted measurements $\hat{\mathbf{y}}^{(i)} = \mathcal{F}(\hat{\mathbf{x}}_1^{(i)})$. We report the mean absolute error (MAE),
\begin{equation}
\label{eq:mae}
\mathrm{MAE}(\mathbf{y})
=
\frac{1}{N}\sum_{i=1}^N \frac{1}{m}\left\|\hat{\mathbf{y}}^{(i)}-\mathbf{y}\right\|_{1},
\end{equation}
and average them across all validation instances.

\paragraph{Case-specific prior/physics fidelity}
Measurement consistency alone is insufficient in scientific inverse problems: a conditional method may match measurements while producing samples that deviate from the learned manifold or violate physics. We therefore evaluate fidelity using metrics tailored to each dataset:

For the 2D toy problems, we form a high-quality reference posterior $p(\cdot\mid\mathbf{y})$ (via importance resampling) and quantify distributional agreement using the 1-Wasserstein distance:
\begin{equation}
\mathcal{W}_1\Big(p(\cdot\mid\mathbf{y}), \hat{p}(\cdot\mid\mathbf{y})\Big)
=
\inf_{\gamma \in \Pi\big(p(\cdot\mid\mathbf{y}),\hat{p}(\cdot\mid\mathbf{y})\big)}
\mathbb{E}_{(\mathbf{x},\hat{\mathbf{x}})\sim\gamma}\left[\|\mathbf{x}-\hat{\mathbf{x}}\|_2\right],
\end{equation}
where $\Pi(\cdot,\cdot)$ denotes the set of all joint distributions, whose marginals are $p(\cdot\mid\mathbf{y})$ and $\hat{p}(\cdot\mid\mathbf{y})$. We complement $\mathcal{W}_1$ with with conditional-sample visualizations to diagnose manifold violations.

For KS dynamics, we measure physics fidelity by the PDE residual computed on generated trajectories using finite differences:
\begin{equation}
R_j^n
=
\frac{u_j^{n+1}-u_j^{n-1}}{2\Delta t}
+
u_j^n\,\frac{u_{j+1}^n-u_{j-1}^n}{2\Delta x}
+
\frac{u_{j+1}^n-2u_j^n+u_{j-1}^n}{(\Delta x)^2}
+
\frac{u_{j+2}^n-4u_{j+1}^n+6u_j^n-4u_{j-1}^n+u_{j-2}^n}{(\Delta x)^4}.
\end{equation}
where $u_j^n \approx u(j\Delta x, n\Delta t)$ is the numerical solution on a discrete grid with spatial step $\Delta x$ and time step $\Delta t$. We summarize physics consistency by the mean absolute residual $\frac{1}{N_xN_t}\sum_{j,n}|R_j^n|$, averaged over conditional samples and validation instances.

For turbulent channel flow dataset, the fidelity is assessed using one-dimensional turbulent kinetic energy (TKE) spectra in the streamwise ($k_x$) and spanwise ($k_z$) directions. Let $\hat{u}_i(k)$ denote the discrete Fourier transform of the $i$-th velocity component along a homogeneous direction; we compute
\begin{equation}
E_{u_i u_i}(k) = k\big\langle \hat{u}_i(k)\hat{u}_i^*(k)\big\rangle,
\end{equation}
where $\langle\cdot\rangle$ denotes averaging over the orthogonal homogeneous direction and over ensemble members. Matching these spectra indicates that conditional samples preserve the characteristic multiscale energy distribution of turbulence beyond pointwise agreement with measurements.

\subsection{Case 1: Toy inverse problems}

We begin with two two-dimensional benchmarks (S-curve and two-moons) as controlled testbeds for conditional sampling. These cases allow (i) direct visualization of conditional distributions and (ii) quantitative comparison against a high-quality reference posterior, making them well suited for diagnosing trade-offs between measurement assimilation and prior fidelity.

\begin{figure}[!htp]
\centering
\includegraphics[width=1.\linewidth]{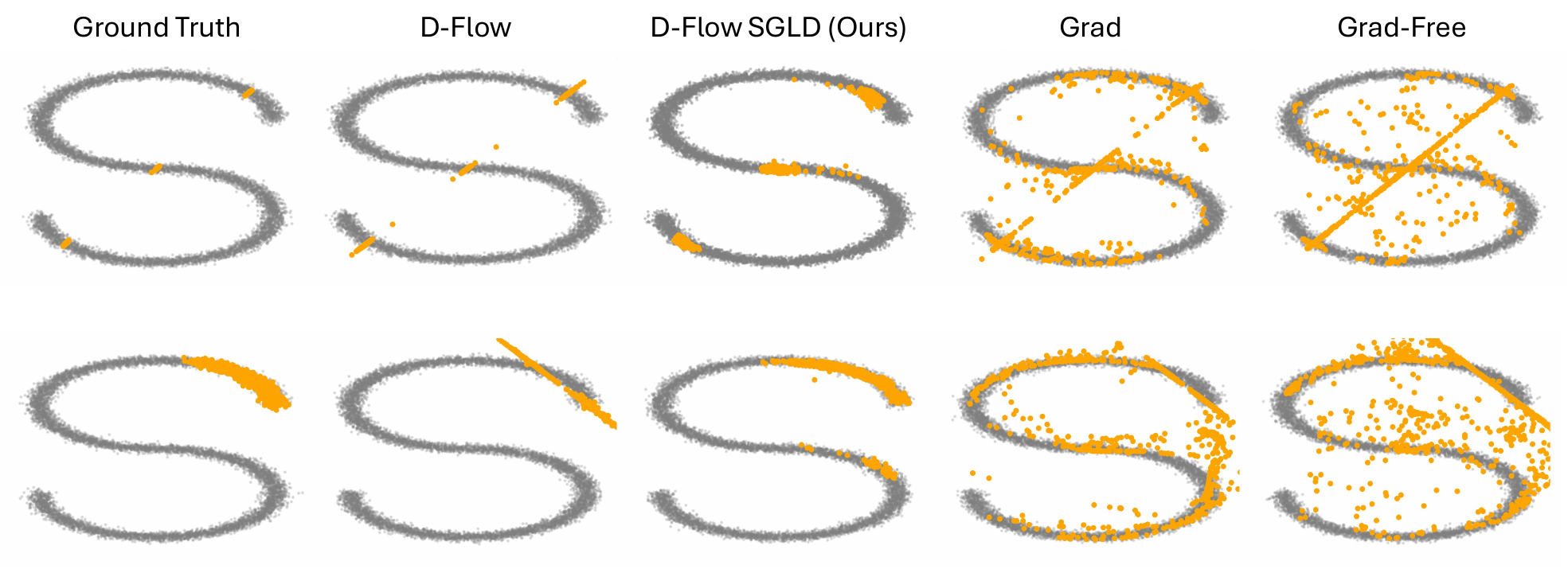}
\caption{S-curve toy inverse problem. Conditional posterior samples (\(N=1000\), orange) overlaid on unconditional OT-CFM samples (gray). For velocity-field guidance, the correction strength is $b = 3$.}
\label{fig:s_curve_res}
\end{figure}

\begin{figure}[!htp]
\centering
\includegraphics[width=1.\linewidth]{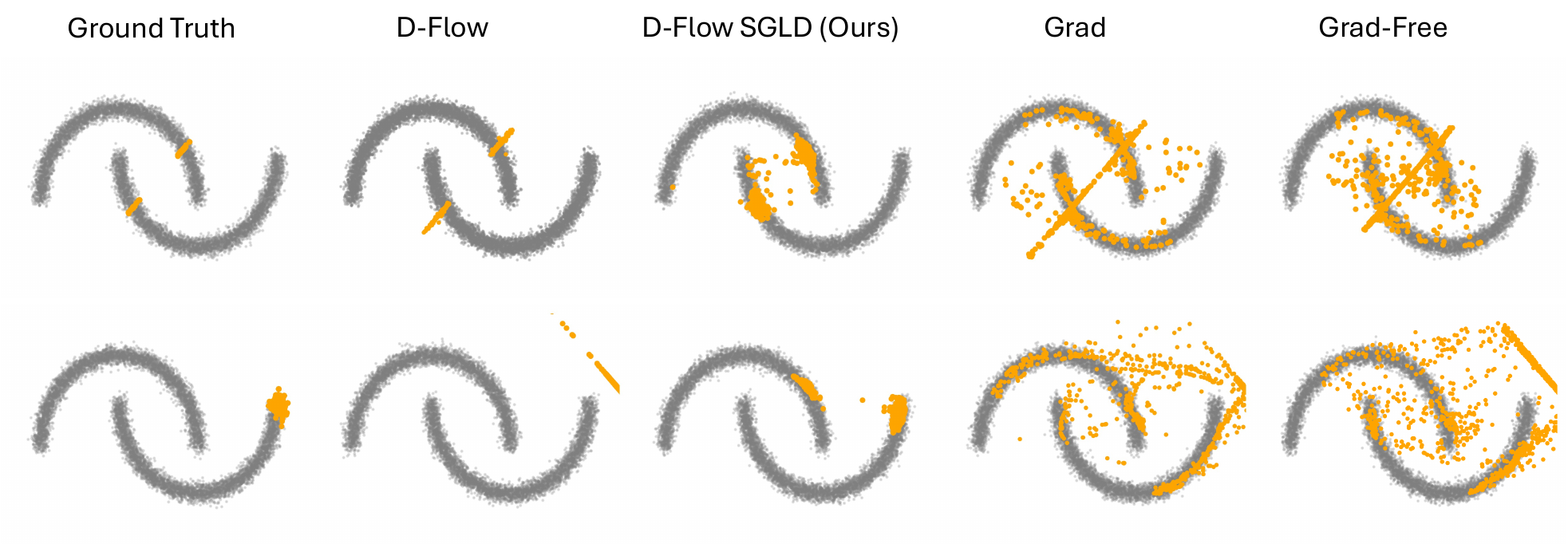}
\caption{Moon-curve toy inverse problem. Conditional posterior samples (\(N=1000\), orange) overlaid on unconditional OT-CFM samples (gray). For velocity-field guidance, the correction strength is $b = 3$}
\label{fig:two_moon_curve_res}
\end{figure}

We consider two observation operators that induce qualitatively different posteriors:
\begin{align*}
\mathcal{F}_1(\mathbf{x}) &= 5|x_0-x_1|,
\qquad
p(y|\mathbf{x}) = \mathcal{N}(\mathcal{F}_1(\mathbf{x}),\,0.01),\\
\mathcal{F}_2(\mathbf{x}) &= x_0+x_1-1.5,
\qquad
p(y|\mathbf{x}) = \mathcal{N}(\mathcal{F}_2(\mathbf{x}),\,0.1).
\end{align*}
Operator $\mathcal{F}_1$ yields a sharply concentrated posterior, resembling scenarios enforcing physical constraints (e.g., conservation laws), whereas \(\mathcal{F}_2\) produces a substantially broader posterior due to higher measurement noise.
For each operator, we form an empirical reference posterior using importance resampling and compare it to conditional samples generated by four training-free methods: velocity-field guidance (Grad / Grad-Free) and source-space inference (D-Flow / D-Flow SGLD). Figures~\ref{fig:s_curve_res} and \ref{fig:two_moon_curve_res} show unconditional samples from the learned prior (gray) and conditional samples (orange) for representative observations.
Across both datasets, the source-space methods (D-Flow and D-Flow SGLD) produce conditional samples that largely remain supported on the learned data manifold, i.e., the orange points mostly lie on or near the gray S-curve/two-moons geometry, with a notable exception for D-Flow when the target posterior occupies the extreme tails of the distribution as seen in the bottom row of Figure~\ref{fig:two_moon_curve_res}. In contrast, velocity-field guidance (Grad and Grad-Free) tends to inject the likelihood signal directly into the sampling dynamics and can enforce measurement consistency at the expense of prior fidelity: a non-negligible fraction of conditional samples visibly depart from the gray manifold and populate off-manifold regions. This effect is particularly apparent when the guidance strength is increased (please refer ~\ref{app:b-comp-study}), indicating that dynamics-level guidance can become overly dominated by the likelihood term and thereby distort the learned distribution. A second qualitative distinction is diversity. D-Flow often yields sharper, more concentrated conditional samples, consistent with deterministic source optimization for each sample. D-Flow SGLD generates a more dispersed set of samples, reflecting stochastic exploration in source space; this can better preserve posterior diversity in broader-posterior settings, but may also introduce low-probability samples that appear as thicker support or weak extra modes relative to the reference.

To quantify distributional accuracy, we compute the 1-Wasserstein distance \(\mathcal{W}_1\) between the empirical conditional distribution and the empirical reference posterior (Table~\ref{table:toy_results}; lower is better).
\begin{table}[htp!]
\centering
\small
\begin{tabular}{lcccc}
\toprule
 & \multicolumn{2}{c}{S-Curve} & \multicolumn{2}{c}{Two-moon} \\
\cmidrule(lr){2-3} \cmidrule(lr){4-5}
Method 
& $\mathcal{F}_1(\cdot)$ 
& $\mathcal{F}_2(\cdot)$ 
& $\mathcal{F}_1(\cdot)$ 
& $\mathcal{F}_2(\cdot)$ \\
\midrule
D-Flow 
& \textbf{0.189} 
& \underline{0.449} 
& \textbf{0.056} 
& \underline{0.251} \\

D-Flow SGLD (Ours) 
& \underline{0.234} 
& \textbf{0.171} 
& \underline{0.073} 
& \textbf{0.160} \\

Grad (b=3) 
& 0.357 
& 0.625 
& 0.129 
& 0.603 \\

Grad-Free (b=3) 
& 0.442 
& 0.596 
& 0.132 
& 0.559 \\
\bottomrule
\end{tabular}
\caption{1-Wasserstein distance \(\mathcal{W}_1\) between the empirical reference posterior (importance resampling) and the empirical conditional distribution for the toy inverse problems (lower is better). Best and second-best results are highlighted in bold and underlined, respectively.}
\label{table:toy_results}
\end{table}
D-Flow achieves optimal performance when the posterior is sharply concentrated, as observed under the measurement likelihood $\mathcal{F}_1(\cdot)$, with D-Flow SGLD ranking second. Conversely, under the measurement likelihood $\mathcal{F}_2(\cdot)$, which induces a more diffuse posterior, alternative approaches exhibit sharp performance degradation. Specifically, guided transport methods (Grad and Grad-Free) and deterministic D-Flow yield substantially higher 1-Wasserstein distances. In these broader posterior regimes, D-Flow SGLD attains the lowest $\mathcal{W}_1$, confirming that stochastic source-space inference provides superior robustness.

Finally, Grad and Grad-Free are sensitive to the guidance strength $b$: as $b$ increases, conditional samples can become progressively more influenced by the likelihood, which may reduce data misfit but also increases off-manifold behavior. We provide a controlled comparison over $b$ in \ref{app:b-comp-study} (Figure~\ref{fig:scale_ablation}), illustrating that guidance-based methods require task-specific tuning to balance assimilation and prior fidelity.

\subsection{Case 2: Kuramoto-Sivashinsky dynamics}
\label{sec:ks_results}

We next evaluate training-free conditioning on solving inverse problems for the one-dimensional KS system, where each sample is a spatiotemporal field $\mathbf{x}\equiv u(x,t)\in\mathbb{R}^{N_x\times N_t}$. We train an OT-CFM prior using a U-Net velocity model $\boldsymbol{\nu}_\theta(\tau,\mathbf{x})$ (architecture details in~\ref{sec:arch-and-hyperparams}). At test time, we condition on partial observations without retraining $\boldsymbol{\nu}_\theta$ and compare four training-free methods: Grad, Grad-Free, D-Flow, and our D-Flow SGLD.

We consider two canonical conditional sampling tasks for chaotic dynamics. The first is one-shot temporal forecasting: given the first \(r\) time frames, we sample the future segment $t>r$, i.e., from $p\big(u(x,t)\mid u(x,t\le r)\big)$. The associated measurement operator extracts the observed prefix,
\begin{equation}
    \mathbf{y} = \mathcal{F}_{1}(u) \equiv u(x,t\le r).
\end{equation}
The second case is to reconstruct the full spatiotemporal field from pointwise measurements at a set of sensor locations, $\mathcal{S}=\{(\hat{x}_\ell,\hat{t}_\ell)\}_{\ell=1}^{|\mathcal{S}|}$, scattered randomly within the space-time domain. The measurement operator, $\mathcal{F}_2$, extracts the sensor measurements from the full field:
\begin{equation}
    \mathbf{y} = \mathcal{F}_{2}(u)\equiv \{u(\hat{x}_\ell,\hat{t}_\ell)\}_{\ell=1}^{|\mathcal{S}|}.
\end{equation}
To probe robustness, we evaluate both \emph{noise-free} observations and observations corrupted by $10\%$ additive noise (relative to the signal scale used in normalization).

\begin{figure}[!htp]
    \centering
    \includegraphics[width=0.9\linewidth]{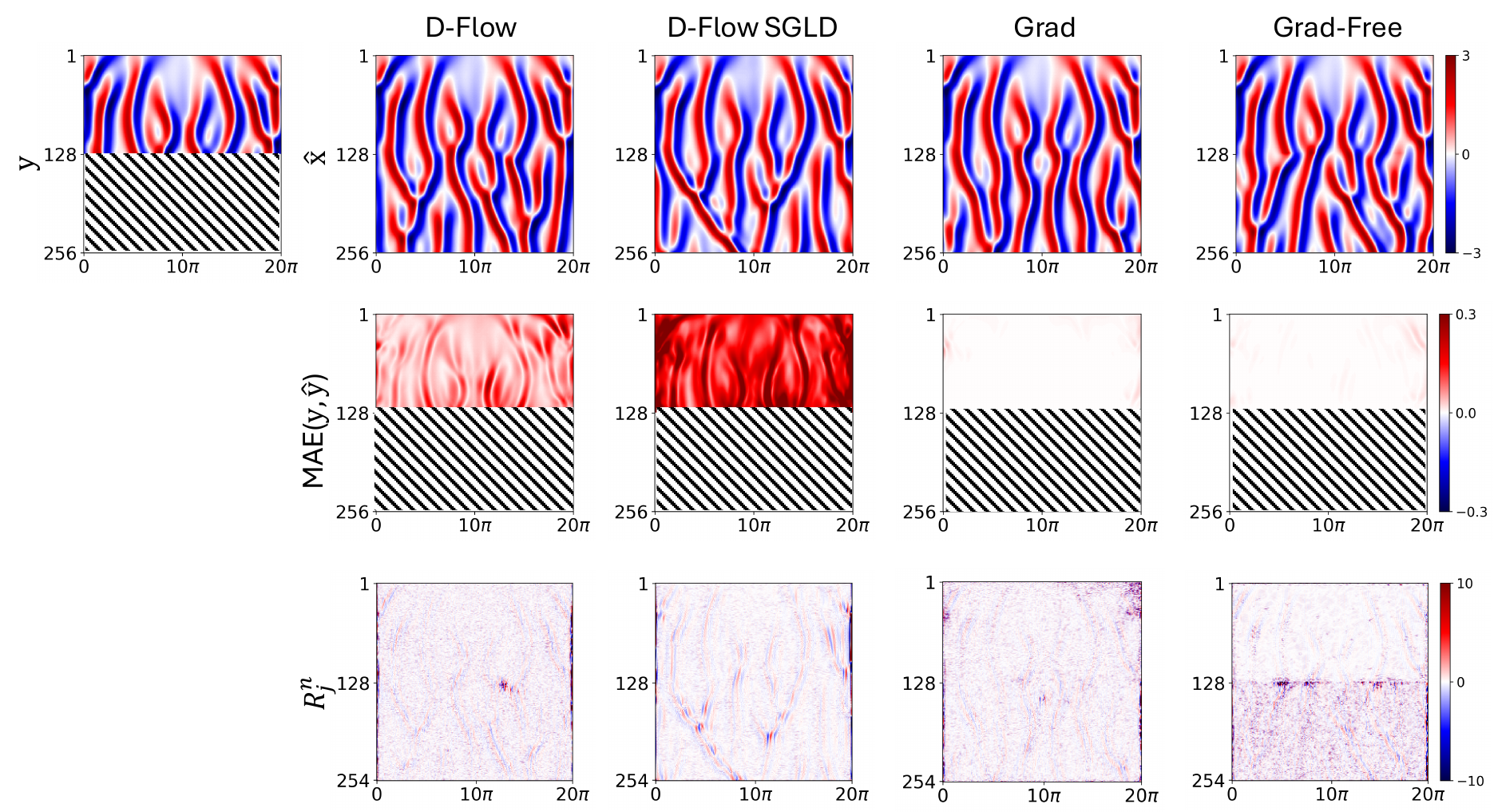}
    \caption{KS temporal forecasting with noise-free measurements. \textbf{(top row)} The ground truth measurement is shown in the first column, followed by a conditionally generated sample from each method. The region below the white line ($t>r$) is the forecasting portion. \textbf{(middle row)} Pointwise data assimilation error (MAE). \textbf{(bottom row)} The PDE residual, $R_j^n$, computed over the entire generated field to assess physical plausibility.}
    \label{fig:KS_contour_noise_free}
\end{figure}

\begin{figure}[!htp]
    \centering
    \includegraphics[width=0.9\linewidth]{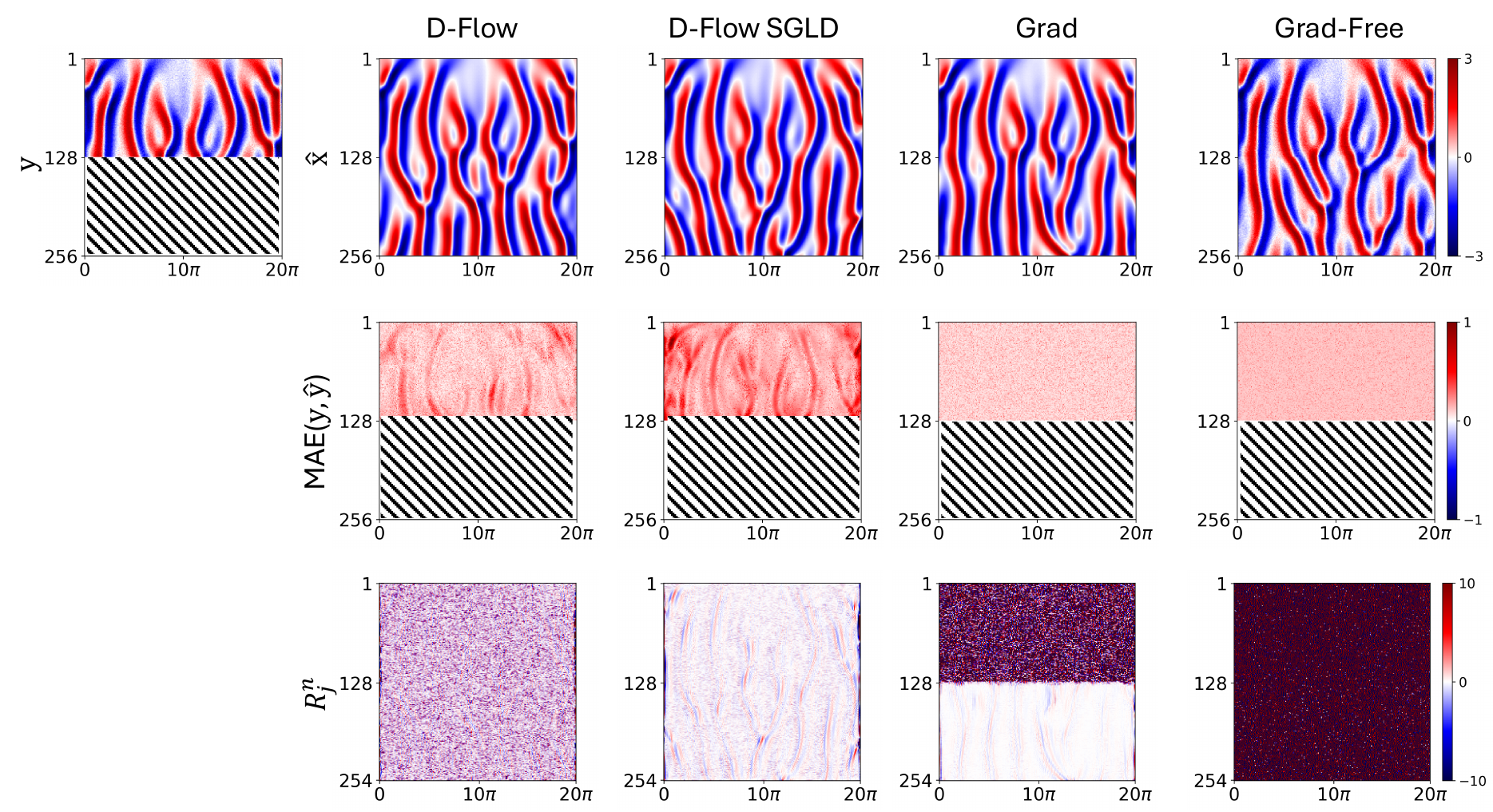}
    \caption{KS temporal forecasting with 10\% noise corruption. \textbf{(top row)} The ground truth measurement is shown in the first column, followed by a conditionally generated sample from each method. The region below the white line ($t>r$) is the forecasting portion. \textbf{(middle row)} Pointwise data assimilation error (MAE). \textbf{(bottom row)} The PDE residual, $R_j^n$, computed over the entire generated field to assess physical plausibility.}
    \label{fig:KS_contour_noise}
\end{figure}

Following Section~\ref{sec:eval-metrics}, we separate \emph{measurement assimilation} (MAE in the observed region) from \emph{prior/physics fidelity} (KS PDE residual). 
Figures~\ref{fig:KS_contour_noise_free} and \ref{fig:KS_contour_noise} visualize temporal forecasting under noise-free and noisy measurements, respectively. The top row shows the measurement (first column) and one representative conditional sample from each method; the white line marks the conditioning boundary $t=r$, and the region $t>r$ is the forecast window. For each method we generate an ensemble of conditional samples given $\mathbf{y}$ and display one randomly selected realization; we generate $100$ samples for Grad, Grad-Free, and D-Flow SGLD, and $20$ samples for D-Flow due to its substantially higher inference cost (Table~\ref{table:comp_costs}).
In the noise-free case (Figure~\ref{fig:KS_contour_noise_free}), all methods produce visually coherent continuations of the observed prefix without obvious seams at the conditioning boundary. Under $10\%$ measurement noise (Figure~\ref{fig:KS_contour_noise}), differences are more pronounced: Grad-Free exhibits a sharp interface near $t=r$, indicating sensitivity to corrupted observations. In contrast, D-Flow and D-Flow SGLD produce smoother continuations that remain visually consistent with the spatiotemporal structures learned by the prior. The middle rows in Figures~\ref{fig:KS_contour_noise_free}-\ref{fig:KS_contour_noise} visualize pointwise MAE on the observed region. As expected for dynamics-level guidance, Grad and (when stable) Grad-Free attain lower MAE than the source-space methods by injecting likelihood information directly into the sampling dynamics, with especially small errors in the noise-free case. The bottom rows assess physics fidelity via the KS residual $R_j^n$: \emph{large residual values indicate violation of the governing KS equation}. In the noise-free setting, all methods yield relatively small residuals, suggesting that the generated trajectories remain close to KS-consistent dynamics. Under noisy measurements, however, Grad and Grad-Free produce markedly larger residuals across the field. Importantly, the residual increase is not confined to the forecast window $t>r$; it is especially elevated in the \emph{observed} region $t\le r$, indicating that the guidance can overfit noise in the measurements in a way that breaks KS consistency even where data are enforced. By contrast, D-Flow and D-Flow SGLD maintain substantially lower residuals in the noisy case, suggesting that source-space inference is more robust to corrupted observations: it avoids forcing the trajectory to match noise. We observe the same qualitative trend in the sparse-sensor reconstruction task; additional results are provided in Section~\ref{sec:KS-sparse-sensor}.

\begin{figure}[t!]
    \centering
    \includegraphics[width=0.85\linewidth]{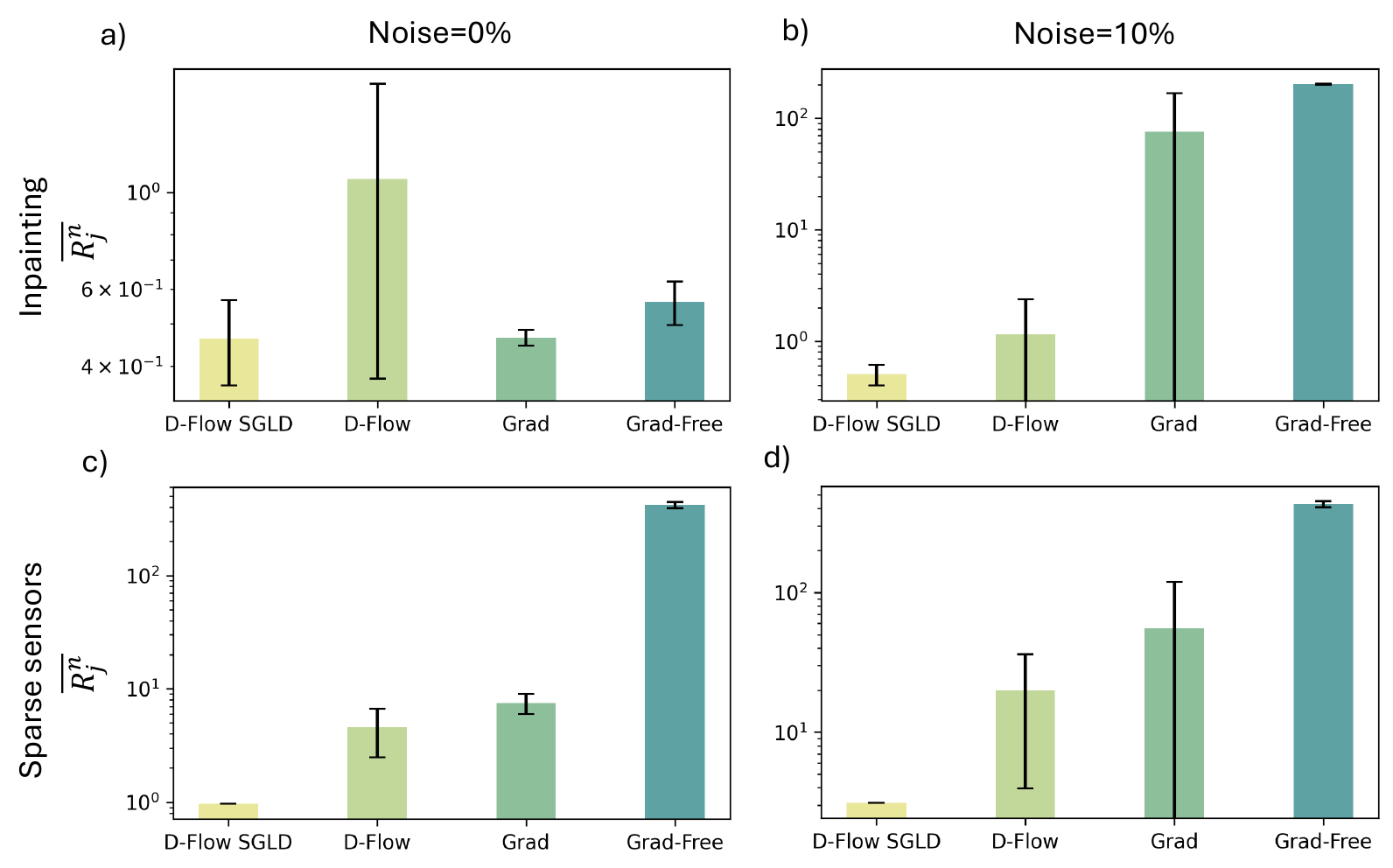}
    \caption{Mean absolute PDE residual, $\overline{R}$, for the different guidance methods applied in KS system. The results are shown for two inverse problems: temporal forecasting (a-b) and reconstruction from sparse sensors (c-d). Each problem is tested with both noise-free measurements (a, c) and measurements corrupted by $10\%$ noise (b, d). Bar heights represent the mean residual, and error bars denote the standard deviation across generated samples.}
    \label{fig:KS_eqn_residual}
\end{figure}

Figure~\ref{fig:KS_eqn_residual} summarizes physics fidelity across validation instances using the mean absolute KS residual, $\overline{R} = \frac{1}{N_t N_x} \sum^{N_t}_{n=1}\sum^{N_x}_{j=1} |R^n_j|$, reported as mean (bar height) and standard deviation (error bar) over generated conditional samples for each validation instance and scenario.
Across both inverse problems (temporal forecasting and sparse-sensor reconstruction) and both noise levels, D-Flow SGLD achieves consistently lowest residuals, indicating robust adherence to the governing KS dynamics even when observations are corrupted. D-Flow achieves comparable mean residuals in several settings but exhibits substantially larger variability in some cases, consistent with sensitivity of deterministic source optimization to the conditioning instance and optimization landscape; moreover, its high inference cost limits practical ensemble generation. Grad and Grad-Free yield moderate residuals for noise-free measurements but degrade sharply under $10\%$ noise, with orders-of-magnitude increases in $\overline{R}$ and, for Grad, pronounced variability (large STD). This behavior is consistent with dynamics-level guidance over-assimilating noisy observations, thereby sacrificing KS consistency and reducing reliability of the resulting conditional samples.


\subsection{Case 3: Wall-bounded Turbulent Flow}

For the final case, we learn an unconditional OT-CFM prior over near-wall velocity fluctuations $\mathrm{x} \equiv \mathbf{u}'$ from a turbulent channel flow at wall-normal location $y^+=40$. Near-wall fluctuations play a central role in the self-sustaining mechanisms of wall-bounded turbulence. Although high-fidelity DNS/LES solvers for wall-bounded turbulence are mature, incorporating sparse and noisy experimental measurements into such simulations remains challenging in practice, especially when one seeks uncertainty-aware reconstructions rather than a single best-fit field. In this case study, we show that a learned OT-CFM prior $\nu_\theta(\tau,\mathbf{x})$, combined with training-free conditioning, provides a practical route to measurement-consistent reconstruction while preserving turbulence-relevant multiscale structure.
To this end, we consider an damaged-data inpainting scenario: a contiguous portion of the spatial field is unobserved and must be imputed with samples that are consistent with the available measurements and remain faithful to turbulent statistics. Since turbulence applications typically require large ensembles for converged statistics, we exclude D-Flow in this case due to its prohibitively high inference cost.

Let $\mathcal{F}$ denote the measurement operator that extracts the observed region of the fluctuation field,
\begin{equation}
    \mathbf{u}'(a_l \leq x \leq a_u,\ b_l \leq z \leq b_u ) = \mathcal{F}(\mathbf{u}'),
\end{equation}
where $a_l, a_u$ and $b_l, b_u$ represent the lower and upper bounds of the available data in the streamwise ($x$) and spanwise ($z$) directions, respectively. This form of partial observation is common in both experiments and simulations. For example, experimental diagnostics often provide measurements only over limited spatial windows or along selected planes, whereas high-fidelity simulation data can be missing due to subsampling, I/O constraints, or data corruption during storage and post-processing. We evaluate both \emph{noise-free} measurements and measurements corrupted with $10\%$ additive noise (relative), using the same conditioning protocol as in the previous cases.

\begin{figure}[htp!]
    \centering
    \includegraphics[width=0.9\linewidth]{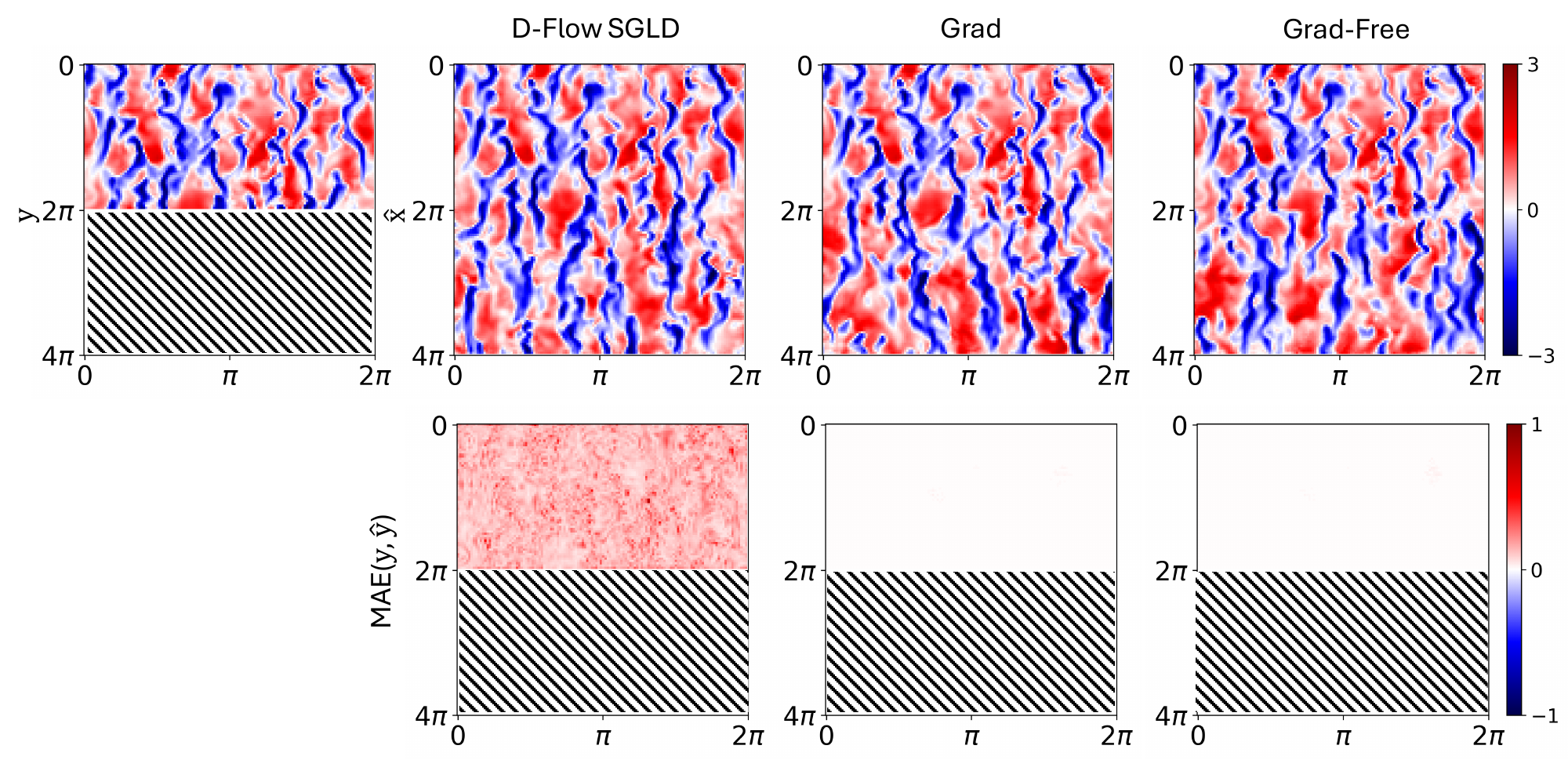}
    \caption{Turbulence inpainting ($u_1$) with noise-free measurements. Top: the observed field (first column) and one representative conditional sample from each method; the hatched region ($x>2\pi$) indicates the missing/inpainted portion. Bottom: pointwise MAE on the observed region.}
    \label{fig:turb_noise_free_contour}
\end{figure}

\begin{figure}[htp!]
    \centering
    \includegraphics[width=0.9\linewidth]{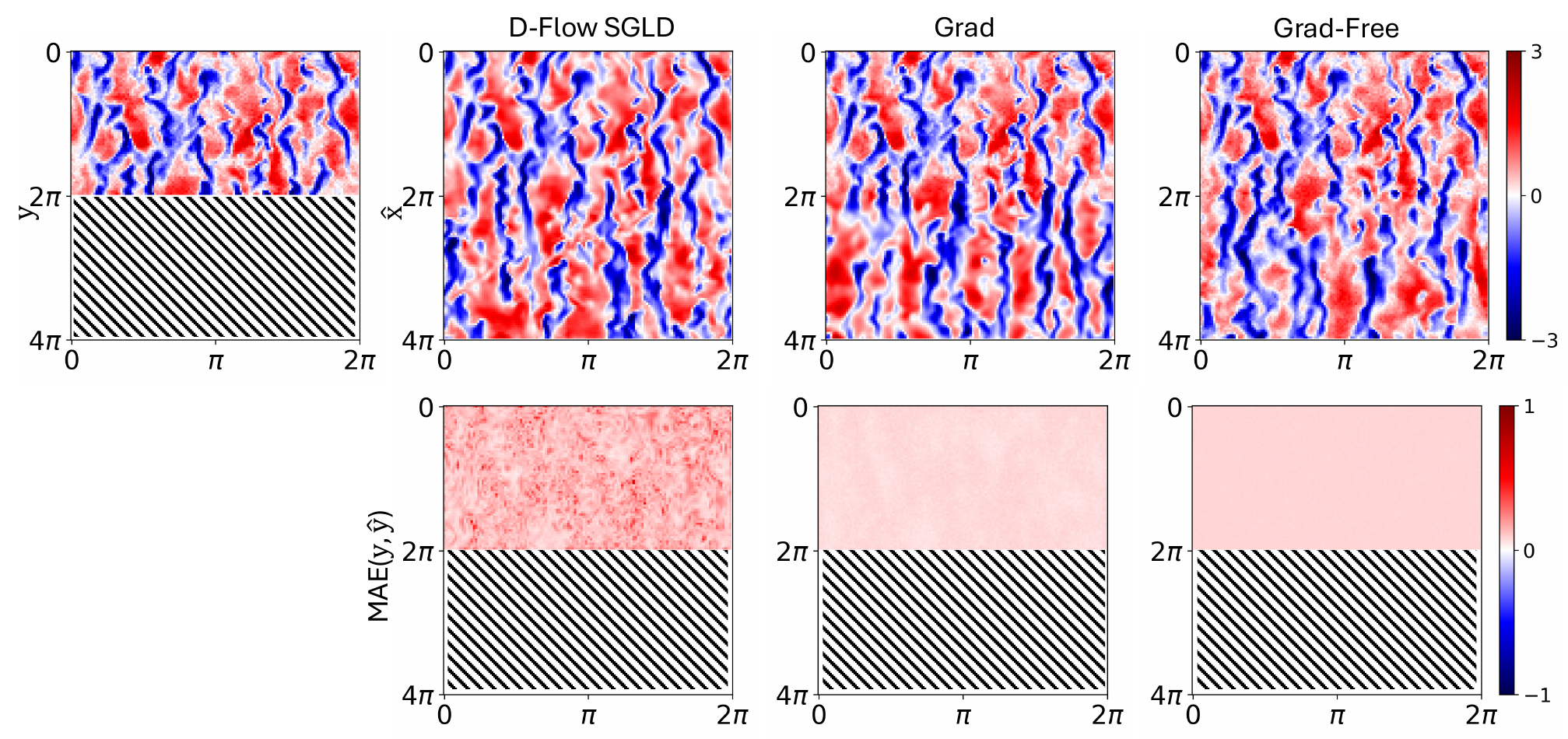}
    \caption{Turbulence inpainting ($u_1$) with noisy measurements ($10\%$). Top: the observed field (first column) and one representative conditional sample from each method; the hatched region ($x>2\pi$) indicates the missing/inpainted portion. Bottom: pointwise MAE on the observed region.}
    \label{fig:turb_noise_contour}
\end{figure}

Figures~\ref{fig:turb_noise_free_contour} and \ref{fig:turb_noise_contour} show representative conditional samples (top row) together with pointwise MAE on the observed region (bottom row), for noise-free and noisy measurements, respectively.
In the noise-free setting, Grad and Grad-Free achieve near-zero MAE on the observed region, indicating very strong measurement assimilation. D-Flow SGLD yields a slightly larger MAE, consistent with stochastic posterior sampling that does not exclusively concentrate mass at the minimum-misfit solutions. Visually, all methods produce inpainted fields that resemble realistic near-wall streaky structures, and the inpainted region does not exhibit an obvious discontinuity at the measurement boundary in the displayed samples.
With $10\%$ noise corruption (Figure~\ref{fig:turb_noise_contour}), the trade-off between assimilation and fidelity becomes apparent. Grad and Grad-Free continue to drive the observed-region MAE close to zero, but the reconstructed fields visibly inherit high-frequency artifacts present in the noisy measurements, suggesting that the guidance can imprint measurement noise into the conditional samples. In contrast, D-Flow SGLD maintains a comparable MAE level to the noise-free case and produces reconstructions that appear less contaminated by noise, indicating improved robustness when the observations are unreliable.

\begin{figure}[!htp]
\centering
\subfloat[Noise-free case]{\includegraphics[width=0.9\textwidth]{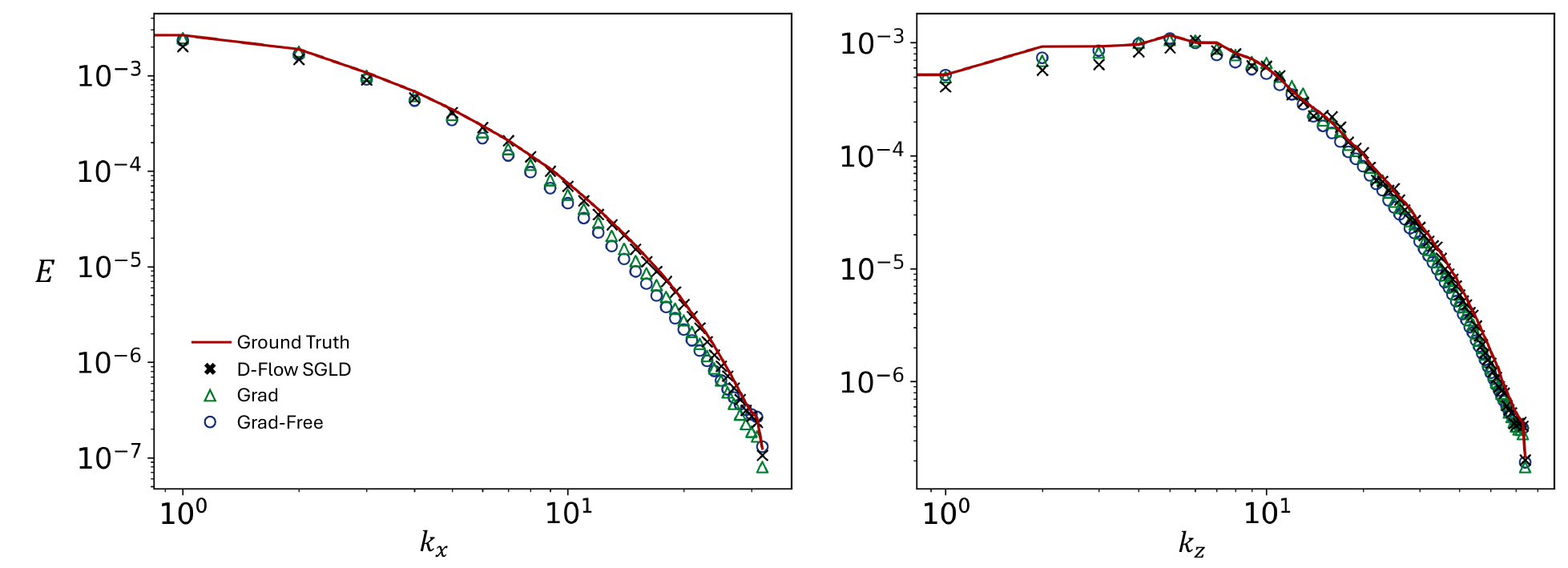}}\\
\subfloat[Noisy, corrupted case] {\includegraphics[width=0.9\textwidth]{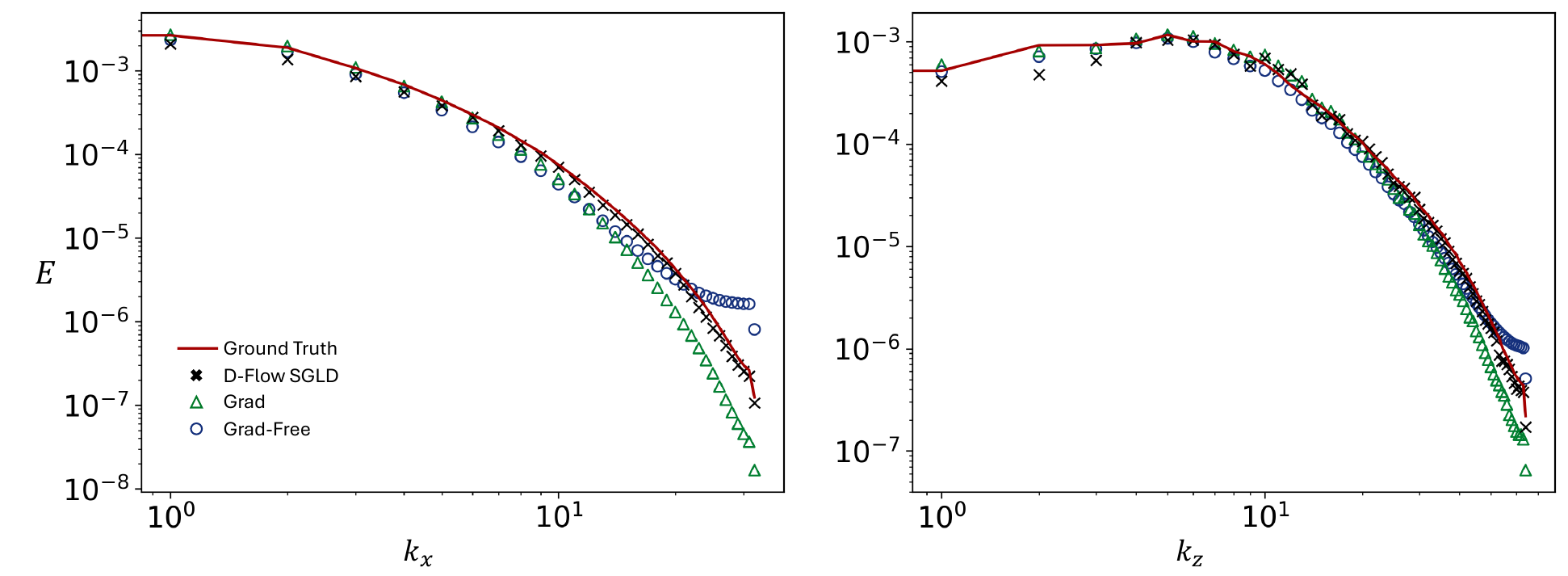}}\\
\caption{Comparison of the streamwise (right) and spanwise (left) one-dimensional energy spectrum for the streamwise velocity component $u_1$ at wall-normal location of $y^+ =40$.}
\label{fig:turb_energy_spectrum}
\end{figure}

Pointwise data misfit does not diagnose whether the reconstruction preserves turbulence physics/statistics across scales. We therefore evaluate physical fidelity using one-dimensional energy spectra in the streamwise $k_x$ and spanwise ($k_z$) directions (Section~\ref{sec:eval-metrics}). Figure~\ref{fig:turb_energy_spectrum} compares the spectra of the streamwise component ($u_1$) against the ground-truth spectra (additional components, $u_i, \ i \in \{2,3\}$, are reported in Section~\ref{sec:additonal-results}). In the noise-free regime (Figure~\ref{fig:turb_energy_spectrum}a), all methods match the ground truth well at low wavenumbers, capturing the large, energy-containing structures. Differences emerge at higher wavenumbers: Grad and Grad-Free exhibit a systematic deficit in high-$k$ energy, indicating loss of small-scale content in the conditional samples, whereas D-Flow SGLD tracks the ground-truth spectrum more closely over the entire wavenumber ranges.
In the noisy measurement scenario (Figure~\ref{fig:turb_energy_spectrum}b), the spectral discrepancies for velocity-field guidance are amplified. The Grad method exhibits a pronounced decay at moderate-to-high wavenumbers, yielding an overly smoothed reconstruction. Grad shows an accelerated roll-off at moderate-to-high wavenumbers, yielding overly smooth reconstructions that underrepresent small-scale fluctuations. Grad-Free exhibits the opposite tendency at the highest wavenumbers, injecting excess high-\(k\) energy consistent with incorporating measurement noise as spurious small-scale content. D-Flow SGLD remains closest to the ground-truth spectrum across both \(k_x\) and \(k_z\), suggesting that stochastic source-space inference better balances measurement information with the learned turbulence prior: it suppresses incoherent noise while retaining coherent multiscale structure.

Overall, the turbulence experiments reinforce the central pattern observed in the KS case: velocity-field guidance (Grad/Grad-Free) can achieve very small data assimilation error but is sensitive to measurement noise and can distort multiscale statistics, while D-Flow SGLD produces more robust conditional samples that better preserve physically relevant spectral structure. 

\section{Discussion}
\label{sec:discussion}

This section provides mechanistic insight into why D-Flow SGLD yields robust conditional samples while remaining computationally scalable. Using a toy problem, we first leverage the reversibility of OT-CFM to visualize how conditional solutions are represented in the source distribution. This perspective reveals a failure mode of deterministic source-space optimization (D-Flow): aggressively enforcing measurement consistency can push solutions into low-density regions of the source prior, which in turn produces off-manifold artifacts after transport to the target space. We then explain how the stochastic source-space sampling dynamics in D-Flow SGLD, together with the source-space prior term defined in our method, mitigates this issue by promoting better-calibrated posterior exploration. Finally, we discuss computational costs and show that D-Flow SGLD substantially reduces the amortized cost of generating large ensembles.


\subsection{Posterior geometry in source space: deterministic optimization vs. stochastic sampling}
\label{subsec:posterior-geometry}

A key characteristic of flow-based models is reversibility. By integrating the negative velocity field, $-\boldsymbol{\nu}_\theta(\tau,\mathbf{x})$, samples from the target distribution $\mathbf{x}_1 \sim p_1(\mathbf{x})$ can be mapped back to the source distribution $\mathbf{x}_0 \sim p_0(\mathbf{x})$. This enables visualization of conditional solutions in the source space: given posterior samples $\mathbf{x}_1 \sim p_1(\mathbf{x}\mid \mathbf{y})$, we compute
\begin{equation}
    \mathbf{x}_0 = \mathbf{x}_1 + \int_{1}^{0} \left(-\nu_\theta(\tau,\mathbf{x}_\tau)\right) d\tau.
\end{equation}
Because OT-CFM learns approximately straight transport paths between $p_0$ and $p_1$, we expect that posterior geometry (e.g., mode separation) is preserved under this inverse mapping.

\begin{figure}[!htp]
    \centering
    \includegraphics[width=0.8\linewidth]{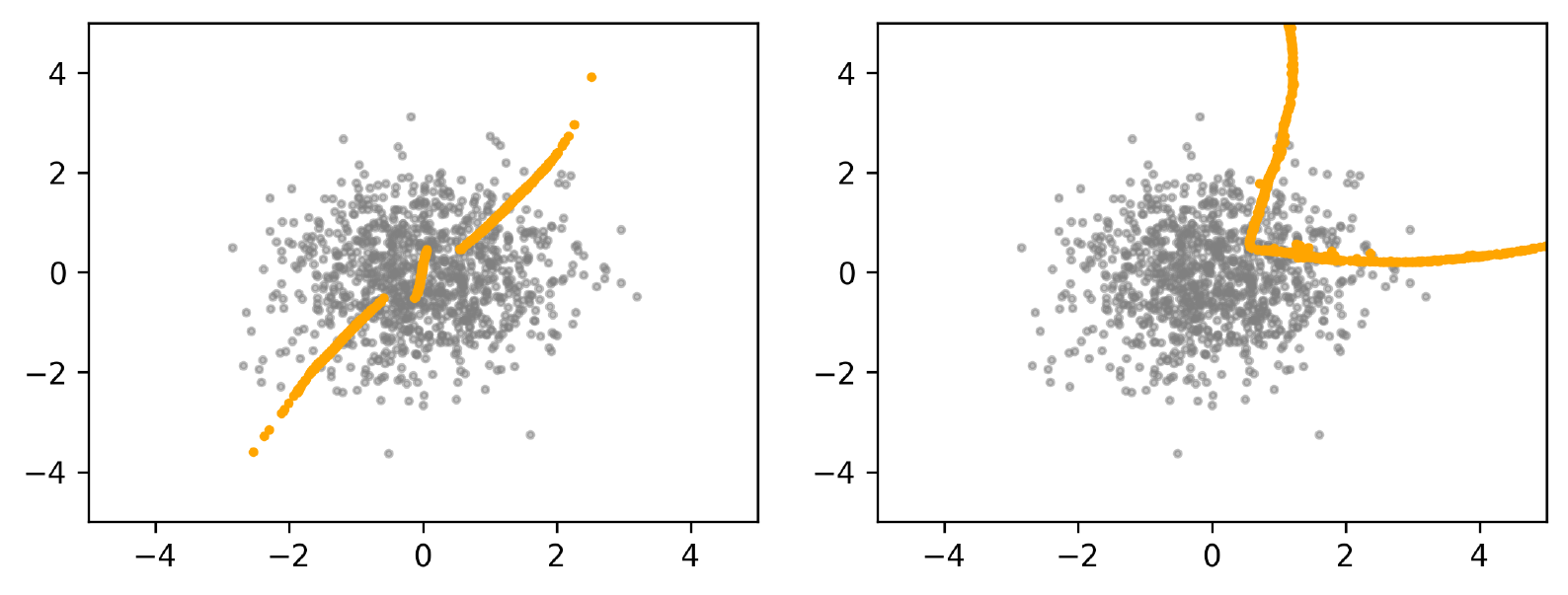}
    \caption{$1000$ source distribution samples obtained after performing optimization using the D-Flow algorithm for S-curve toy problem for the measurement functions $\mathcal{F}_1(\cdot)$ (left) and $\mathcal{F}_2(\cdot)$ (right) respectively. The gray dots represent the samples for the unconditional Gaussian distribution, while the orange samples represents the samples generated after optimizing the source terms using the D-Flow Algorithm.}
    \label{fig:d-flow-source}
\end{figure}

Figure~\ref{fig:d-flow-source} supports this hypothesis: distinct modes in the conditional distribution remain separated when mapped back to the source. At the same time, Figure~\ref{fig:d-flow-source} reveals an important limitation of deterministic source-space optimization (D-Flow).
Although D-Flow can produce diverse conditional outputs by initializing from different random seeds (yielding a set of independently optimized source points), each sample is obtained via a deterministic optimization trajectory that strongly prioritizes measurement satisfaction. As a result, some optimized source points are driven into low-probability regions under the source prior (i.e., far from the central Gaussian). After forward transport, these low-density source configurations can manifest as off-manifold or degraded samples in the target space, consistent with the failure modes observed in Figures~\ref{fig:s_curve_res} and~\ref{fig:two_moon_curve_res}. This source-space view therefore provides a concrete geometric explanation for why deterministic optimization can be brittle under noisy or partially informative measurements: measurement overfitting may be achieved by leaving the high-density region of the learned prior.

This observation also clarifies the role of the source-space prior term used in our D-Flow SGLD. Specifically, D-Flow SGLD targets a regularized objective that balances measurement consistency with source prior plausibility:
\begin{equation}
    \min_{\mathbf{x}_0}\ \ \|\mathbf{y}-\mathcal{F}(\mathbf{x}_1(\mathbf{x}_0))\|_2^2
    + \lambda\|\mathbf{x}_0\|_2^2,
\end{equation}
where $\mathbf{x}_1(\mathbf{x}_0)$ denotes the forward transport of $\mathbf{x}_0$ under the learned velocity field $\boldsymbol{\nu}_\theta$, and $\lambda$ controls the strength of the source-space prior penalty. Importantly, unlike multi-start deterministic optimization, D-Flow SGLD introduces stochastic source-space dynamics (via Langevin updates) that enable exploration of a neighborhood of high-probability source configurations rather than committing to a single optimized point per run.

\begin{figure}[!htp]
    \centering
    \includegraphics[width=0.8\linewidth]{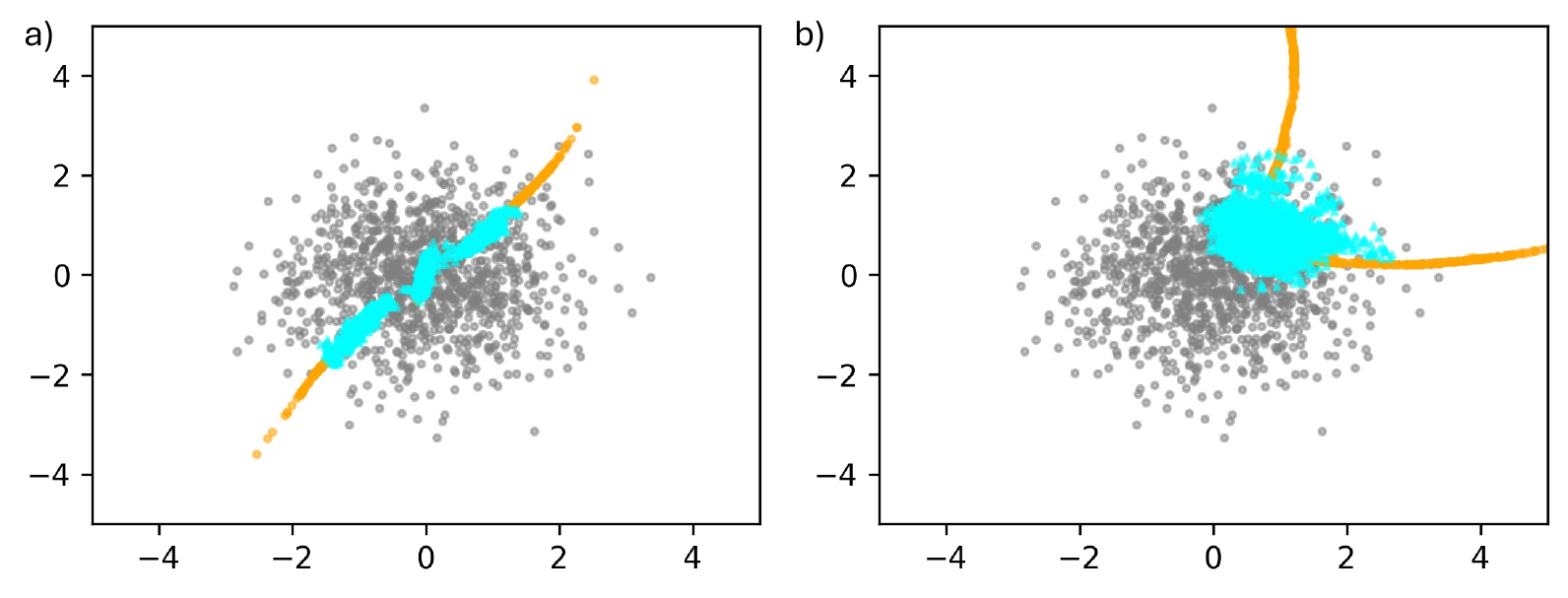}
    \caption{$1000$ source distribution samples obtained after performing optimization using the D-Flow SGLD algorithm for S-curve toy problem for the measurement functions $\mathcal{F}_1(\cdot)$ (left) and $\mathcal{F}_2(\cdot)$ (right) respectively. The gray dots represent the samples for the unconditional Gaussian distribution. The orange dots represents the samples generated after optimization using the D-Flow Algorithm. The cyan dots represents the samples generated with source term regularization using the D-Flow SGLD Algorithm.}
    \label{fig:d-flow-sgld-source}
\end{figure}

Figure~\ref{fig:d-flow-sgld-source} illustrates the effect of this mechanism. 
Without enforcing source-space plausibility, optimization can push conditional solutions far outside the typical set of $p_0$. With the source prior term and stochastic exploration, conditional samples remain largely within the high-density region of the source while still reflecting the posterior structure induced by the measurements. This yields better-calibrated posterior exploration and mitigates off-manifold generations, providing a direct explanation for the robustness of D-Flow SGLD under adverse conditions.

\subsection{Computational Costs}

Most training-free guidance methods map a set of initial source points to the same number of target samples, making the cost scale linearly with the desired ensemble size. This becomes prohibitive when large ensembles are needed to estimate converged posterior statistics, as in turbulent flows. Deterministic source-space optimization (D-Flow) is especially expensive in this regime because each posterior sample requires a separate optimization solve.

D-Flow SGLD offers a more cost-effective alternative. As an MCMC-based approach, once burn-in is reached, a single chain can generate many correlated samples at low incremental cost, reducing the amortized cost per posterior sample relative to independent optimization-based sampling. Gemini said
Table~\ref{table:comp_costs} summarizes the wall-clock inference costs, utilizing a single NVIDIA L40 GPU for the toy problems and a single NVIDIA A100 GPU for the higher-dimensional KS and turbulence cases. These metrics explicitly highlight the operational trade-off between computational speed and methodological robustness. Grad-Free is the fastest but is less accurate and more sensitive to measurement noise. D-Flow is accurate but incurs prohibitive costs for large-scale problems. Grad remains accurate but can still be expensive. D-Flow SGLD strikes a favorable balance by producing robust and physically plausible conditional samples at a cost that is substantially lower than D-Flow and competitive ($\sim2\times$ speedup) to gradient-based guidance, making it a practical strategy for large-ensemble conditional generation in complex scientific and engineering settings.
\begin{table}[t]
\centering
\small
\begin{tabular}{lcccc}
\toprule
Case & D-Flow & D-Flow SGLD & Grad & Grad-Free \\
\midrule
Toy ($N=1000$)   & 65   & 8    & 8    & 5   \\
KS ($N=100$)    &  $12000^*$ &  578 & 1114  &  363 \\
Turb ($N=1000$)  & $200500^\dagger$ &  928 & 1841 & 840 \\

\bottomrule
\end{tabular}
\caption{Wall-clock inference time (seconds) for training-free conditional generation methods on a single NVIDIA L40 GPU for the toy case and a single NVIDIA A100 80GB GPU for the rest of the cases. $N$ denotes the number of posterior samples generated. $\*$ and $\dagger$ denotes extrapolation of the cost by generating 20 and 1 sample(s) respectively.}
\label{table:comp_costs}
\end{table}

\section{Conclusion}

We studied training-free conditional generation for scientific inverse problems using pretrained FM priors, with emphasis on posterior sampling that remains faithful to learned physical structure under partial and noisy observations. We organized inference-time conditioning strategies into two families: (i) velocity-field guidance, which injects likelihood information by perturbing the transport dynamics during sampling, and (ii) source-space inference, which keeps the learned transport map fixed and instead performs posterior inference over the source variable whose pushforward yields measurement-consistent states. Building on the source-space perspective, we introduced D-Flow SGLD, a preconditioned SGLD sampler targeting the induced source posterior, enabling uncertainty-aware conditional sampling without retraining the FM prior.

Across a hierarchy of benchmarks, including 2D toy posteriors, chaotic KS dynamics, and wall-bounded turbulent flow reconstruction, we observed a consistent assimilation-fidelity trade-off. Dynamics-level guidance methods can achieve very small assimilation error but are sensitive to guidance strength and can overfit corrupted observations, leading to degraded physics consistency (e.g., elevated KS residuals) and distorted multiscale turbulence statistics (e.g., spectral bias at high wavenumbers). Deterministic source optimization (D-Flow) can yield accurate conditional reconstructions but is computationally expensive for large ensembles and can exhibit instance-dependent variability. In contrast, D-Flow SGLD provides a robust compromise: by sampling in source space under an explicit prior term, it better avoids tail-seeking behavior that produces off-manifold artifacts, maintains physics/statistics fidelity under noise, and reduces the amortized cost of producing the large conditional ensembles needed for scientific uncertainty quantification.

These results highlight source-space posterior sampling as a practical and scalable conditioning mechanism for FM priors in scientific inverse problems, and motivate future work on improved mixing diagnostics (e.g., effective sample size), richer source-space regularization, and extensions to more complex observation operators and higher-Reynolds-number turbulent flows.

\section*{Acknowledgment}
\noindent The authors would like to acknowledge the funds from the Office of Naval Research (Award No. N00014-23-1-2071) and National Science Foundation (Award No. OAC-2047127).

\section*{Declaration of Generative AI use}

\noindent During the preparation of this work, the authors used Large Language Models like ChatGPT etc., to polish the grammar and improve the readability of the manuscript.

\clearpage

\appendix

\section{Connection to score-based posterior guidance}
\label{app:score_flow_connection}

This appendix provides the derivation underlying the velocity-field guidance used in
Section~\ref{subsec:vel_guidance}. We first recall the probability-flow ODE formulation of score-based diffusion models~\cite{song2020score}, then show how Bayes' rule induces an additive likelihood-driven correction when targeting a posterior $p(\mathbf{x}_1|\mathbf{y})$. Finally, we summarize the DPS approximation~\cite{chung2022diffusion,chung2025diffusion} and explain how it connects to the plug-in likelihood-gradient proxy used in our OT-CFM guided transport.

Consider a forward diffusion process that transforms clean data $\mathbf{x}_1\sim p_1$ into a noisy variable
$\mathbf{x}_\tau$ by the It\^{o} SDE
\begin{equation}
\label{eq:app_forward_sde}
d\mathbf{x} = f(\tau,\mathbf{x}_\tau)d\tau + g(\tau)d\mathbf{w}_\tau,
\end{equation}
where $\mathbf{w}_\tau$ is standard Brownian motion. Under standard regularity conditions, the time-reversed process is also an SDE with drift involving the score $\nabla_{\mathbf{x}}\log p_\tau(\mathbf{x}_\tau)$ \cite{song2020score}:
\begin{equation}
\label{eq:app_reverse_sde}
d\mathbf{x} =
\big[f(\tau,\mathbf{x}_\tau) - g(\tau)^2 \nabla_{\mathbf{x}}\log p_\tau(\mathbf{x}_\tau)\big]d\tau
+ g(\tau)d\widehat{\mathbf{w}}_\tau,
\end{equation}
where $\widehat{\mathbf{w}}_\tau$ denotes reverse-time Brownian motion. A deterministic sampler with the same marginal densities $\{p_\tau\}_{\tau\in[0,1]}$ is given by the probability-flow ODE (PF-ODE) \cite{song2020score}
\begin{equation}
\label{eq:app_pfode}
\frac{d\mathbf{x}_\tau}{d\tau}
=
f(\tau,\mathbf{x}_\tau) - \frac{1}{2}g(\tau)^2 \nabla_{\mathbf{x}}\log p_\tau(\mathbf{x}_\tau).
\end{equation}
It is clear that the modified drift term (RHS) of PF-ODE is the transport velocity in CFM. 

Let $\mathbf{y}$ denote observations generated by a likelihood model $p(\mathbf{y}|\mathbf{x}_1)$ (e.g., the
measurement equation in Section~\ref{subsec:general_inverse}). Bayes' rule implies
$p(\mathbf{x}_1|\mathbf{y}) \propto p(\mathbf{x}_1)p(\mathbf{y}|\mathbf{x}_1)$, hence
\begin{equation}
\label{eq:app_bayes_score}
\nabla_{\mathbf{x}_1}\log p(\mathbf{x}_1|\mathbf{y}) = \nabla_{\mathbf{x}_1}\log p(\mathbf{x}_1) + \nabla_{\mathbf{x}_1}\log p(\mathbf{y}|\mathbf{x}_1).
\end{equation}
In diffusion posterior sampling, an analogous identity holds at intermediate times, yielding a posterior score at $\tau$ of the form $\nabla_{\mathbf{x}_\tau}\log p_\tau(\mathbf{x}_\tau|\mathbf{y}) = \nabla_{\mathbf{x}_\tau}\log p_\tau(\mathbf{x}_\tau) + \nabla_{\mathbf{x}_\tau}\log p_\tau(\mathbf{y}|\mathbf{x}_\tau)$.

Substituting the posterior score decomposition into the PF-ODE yields a posterior probability-flow ODE:
\begin{align}
\frac{d\mathbf{x}_\tau}{d\tau}
&=
f(\tau,\mathbf{x}_\tau) - \frac{1}{2}g(\tau)^2 \nabla_{\mathbf{x}}\log p_\tau(\mathbf{x}_\tau|\mathbf{y}) \nonumber\\
&=
\underbrace{\Big[f(\tau,\mathbf{x}_\tau) - \frac{1}{2}g(\tau)^2 \nabla_{\mathbf{x}}\log p_\tau(\mathbf{x}_\tau)\Big]}_{\text{prior PF-ODE drift (velocity)}}
\;-\;
\underbrace{\frac{1}{2}g(\tau)^2 \nabla_{\mathbf{x}}\log p_\tau(\mathbf{y}|\mathbf{x}_\tau)}_{\text{likelihood-driven correction}}.
\label{eq:app_pfode_posterior}
\end{align}
Thus, relative to the unconditional (prior) sampler, conditioning on $\mathbf{y}$ introduces an additive correction term proportional to the likelihood score. This is the conceptual basis for posterior score guidance in diffusion models~\cite{song2020score,chung2022diffusion,chung2025diffusion}, as well as the velocity-field guidance in CFM models. 

\subsection{DPS approximation and plug-in likelihood gradients}
\label{app:dps}

A key practical challenge is that $\nabla_{\mathbf{x}}\log p_\tau(\mathbf{y}\mid\mathbf{x}_\tau)$ is typically
intractable because $\mathbf{x}_\tau$ is a noisy/intermediate variable whereas the measurement operator $\mathcal{F}$
is defined on the clean state $\mathbf{x}_1$. Diffusion posterior sampling (DPS) \cite{chung2022diffusion,chung2025diffusion}
approximates the likelihood score by (i) constructing a denoised predictor $\widehat{\mathbf{x}}_{1\mid\tau}$ from
$\mathbf{x}_\tau$, and (ii) using a plug-in gradient:
\begin{equation}
\label{eq:app_dps_proxy}
\nabla_{\mathbf{x}_\tau}\log p_\tau(\mathbf{y}\mid\mathbf{x}_\tau)
\;\approx\;
\nabla_{\mathbf{x}_\tau}\log p(\mathbf{y}\mid \widehat{\mathbf{x}}_{1\mid\tau}).
\end{equation}
For Gaussian measurement noise $p(\mathbf{y}\mid \mathbf{x}_1)=\mathcal{N}(\mathcal{F}(\mathbf{x}_1),\sigma_y^2\mathbf{I})$,
the plug-in gradient becomes
\begin{equation}
\label{eq:app_dps_gaussian}
\nabla_{\mathbf{x}_\tau}\log p(\mathbf{y}\mid \widehat{\mathbf{x}}_{1\mid\tau})
=
-\frac{1}{\sigma_y^2}\nabla_{\mathbf{x}_\tau}\left\|\mathbf{y}-\mathcal{F}\left(\widehat{\mathbf{x}}_{1\mid\tau}\right)\right\|_2^2,
\end{equation}
where gradients are computed by AD through $\mathcal{F}$ and through the predictor
$\widehat{\mathbf{x}}_{1\mid\tau}$.

\subsection{From diffusion guidance to flow-matching guided transport}
\label{app:flow_guidance}

Flow matching learns a transport velocity field $\nu_\theta(\tau,\mathbf{x})$ that maps $p_0$ to $p_1$ via the ODE
\eqref{eq:transport_general} without specifying $(f,g)$ explicitly. Nevertheless, motivated by the additive correction
structure in \eqref{eq:app_pfode_posterior}, we adopt the guided transport form
\begin{equation}
\label{eq:app_guided_transport}
\frac{d\mathbf{x}_\tau}{d\tau}
=
\nu_\theta(\tau,\mathbf{x}_\tau) + \lambda(\tau)\mathbf{g}(\tau,\mathbf{x}_\tau;\mathbf{y}),
\end{equation}
where $\mathbf{g}$ is chosen to align with the DPS plug-in likelihood gradient \eqref{eq:app_dps_gaussian}.

For OT-CFM, the bridge construction \eqref{eq:otcfm_bridge} implies near-straight trajectories, motivating the
straight-path predictor used in the main text:
\begin{equation}
\label{eq:app_otcfm_predictor}
\widehat{\mathbf{x}}_{1\mid\tau} = \mathbf{x}_\tau + (1-\tau)\nu_\theta(\tau,\mathbf{x}_\tau).
\end{equation}
Substituting \eqref{eq:app_otcfm_predictor} into \eqref{eq:app_dps_gaussian} yields the proxy gradient used in
Section~\ref{subsec:vel_guidance}.

\section{Algorithm for Preconditioned D-Flow SGLD}
\label{sec:d-flow-sgld-algo}

The D-Flow SGLD method, summarized in Algorithm~\ref{alg:d-flow-sgld}, is a parallel MCMC-based approach for sampling from the posterior distribution by iteratively updating a sample in the source (latent) space, $\mathrm{x}_0$. Each iteration consists of four main steps.

\begin{algorithm}[H]
\caption{D-Flow SGLD}\label{alg:d-flow-sgld}
\small
\SetAlgoLined
Require: $\nu_\theta$, $\mathrm{N_{parallel}}$, $\mathrm{N_{steps}}$, $\mathrm{burn}$, $\mathrm{x^{(0)}_0}$, $\mathrm{y}$, $\mathcal{F}$, $\mathcal{R}$, $\lambda$, $\omega$, $\eta$, $\delta$, $s$\;
collect = [\ ]\;
\SetKwBlock{Parallel}{do in parallel for $1:\mathrm{N_{parallel}}$}{end}
\Parallel{
\For{$i \gets 1 : \mathrm{N_{steps}}$}{
    $\mathrm{x}^{(i-1)}_1 \gets \mathrm{x}^{(i-1)}_0 + \int^1_{\tau=0} \nu_\theta(\tau, \mathrm{x}^{(i-1)}_\tau)d\tau$\;
    {$\mathrm{L}_1 \gets \|\mathrm{y} -  \mathcal{F}(\mathrm{x}^{(i-1)}_1) \|^2_2$\;
    \uIf{use regularizer}
{        $\mathrm{L}_2 \gets  \lambda \mathcal{R}(\mathrm{x}^{(i-1)}_0)$\;
        $\mathrm{L} = \mathrm{L}_1 + \mathrm{L}_2$\;}}
    \Else{$\mathrm{L} = \mathrm{L}_1$\;}
    \uIf{use preconditioner}{
       $ V \gets \omega V + (1-\omega) \nabla_{{x}^{(i-1)}_0} \mathrm{L}_1 \odot \nabla_{{x}^{(i-1)}_0} \mathrm{L}_1$\;
       $ \mathbf{P}_k \gets diag(1\ \oslash \ (\sqrt{V} + \delta(i))) $\;
    }
    \Else{
        $\mathbf{P}_k \gets \mathbf{I}$\;
    }
    $\mathrm{x}^{(i)}_0 \gets \mathrm{x}^{(i-1)}_0 -\  \eta(i)\ \mathbf{P}_k\  \nabla_{{x}^{(i-1)}_0} \mathrm{L} + \mathcal{N}(0,\ 2\eta(i)s(i)\mathbf{P}_k) $\;
    \uIf{$i \geq \mathrm{burn}$}{
        collect.append($\mathrm{x}^{(i)}_0$)\;
    }
}
}
return collect\;
\end{algorithm}

First, a forward pass maps the current latent sample $\mathrm{x}_0^{(i-1)}$ to the target-space sample $\mathrm{x}_1^{(i-1)}$ using the pre-trained flow model $\nu_\theta$. Second, a loss function, $L$, is computed, which includes a data-consistency term based on the measurement $\mathrm{y}$ and an optional source-space regularization term, $R(\mathrm{x}_0)$. Third, an optional adaptive preconditioner, $\mathbf{P}_k$, is calculated to scale the gradients for high-dimensional problems. Finally, the latent sample is updated using a preconditioned Langevin Dynamics step, which follows the gradient of the loss and incorporates annealed noise to guide the exploration. Samples of $\mathrm{x}_0$ are collected after a specified burn-in period to ensure they represent the target posterior distribution.

\section{Toy data generation details}
\label{app:toy_data}

This appendix specifies the exact sampling procedures used to generate the 2D toy datasets.

\paragraph{S-curve}
We sample a latent parameter $t \sim \mathrm{Unif}\!\left(-\frac{3\pi}{2},\,\frac{3\pi}{2}\right)$ and define
\begin{equation}
\mathbf{x}
=
\begin{bmatrix}
\sin(t)\\
0.5\,\mathrm{sgn}(t)\,(\cos(t)-1)
\end{bmatrix}
+\epsilon,
\qquad
\epsilon \sim \mathcal{N}(0,\sigma^2\mathbf{I}).
\end{equation}

\paragraph{Two-moons}
Given a requested batch size $B$, we split samples into two semicircles:
$N_{\mathrm{out}}=\lfloor B/2\rfloor$ and $N_{\mathrm{in}}=B-N_{\mathrm{out}}$.
For each subset we draw $\phi \sim \mathrm{Unif}(0,\pi)$ and set
\begin{equation}
\mathbf{x}_{\mathrm{base}}=
\begin{cases}
\begin{bmatrix}
\cos(\phi)\\
\sin(\phi)
\end{bmatrix}, & \text{for } N_{\mathrm{out}} \text{ samples},\\[0.8em]
\begin{bmatrix}
1-\cos(\phi)\\
0.5-\sin(\phi)
\end{bmatrix}, & \text{for } N_{\mathrm{in}} \text{ samples}.
\end{cases}
\end{equation}
We then apply isotropic noise and an affine rescaling:
\begin{equation}
\mathbf{x}=
\frac{1}{1.5}\left(
\mathbf{x}_{\mathrm{base}}+\epsilon+
\begin{bmatrix}
-0.5\\
-0.25
\end{bmatrix}
\right),
\qquad
\epsilon \sim \mathcal{N}(0,\sigma^2\mathbf{I}).
\end{equation}

\section{Details about architecture, training, and inference hyperparameters}
\label{sec:arch-and-hyperparams}

For the KS equation and wall-bounded turbulence cases, we parameterize the flow-matching velocity field $\boldsymbol{\nu}_\theta(\tau,\mathbf{x}_\tau)$ using a modernized U-Net~\cite{ronneberger2015u}. The architecture follows a hierarchical multi-resolution design with residual blocks in both the downsampling and upsampling paths. Each block uses SiLU activations and convolutional layers; self-attention is included at selected resolutions when enabled. The transport time \(\tau\) is mapped to a high-dimensional positional embedding, processed by an MLP, and injected into the network via feature-wise linear modulation (FiLM)~\cite{perez2018film} to condition the velocity prediction on \(\tau\).
We train $\boldsymbol{\nu}_\theta$ using the OT-CFM objective. Mini-batch optimal transport is used to construct source--target sample couplings during training, using the OT solver implementation provided by the \texttt{torchcfm} repository. The OT-CFM path parameter $\sigma$ is set to \(10^{-3}\). Model parameters are optimized with Adam using a fixed learning rate of \(10^{-4}\).

Training-free conditional generation additionally depends on inference-time hyperparameters (e.g., guidance strength, step sizes, and sampler settings). Table~\ref{tab:hyperparams} summarizes the hyperparameters used for each method and dataset in the main experiments.

\begin{table}[H]
\centering
\caption{Hyperparameters for Training-Free Guidance Methods Across Different Cases.}
\label{tab:hyperparams}
\scriptsize
\setlength{\tabcolsep}{4pt}
\renewcommand{\arraystretch}{1.5} 
\begin{tabular}{l l l l l}
\toprule
\textbf{Case} & \textbf{D-Flow} & \textbf{D-Flow SGLD} & \textbf{Grad/Grad-Free}\\
\midrule
\textbf{Toy} & 
\makecell[tl]{ 
    $N_{optim}= {10\ \text{(S-curve)},\ 20\ \text{(Two-moon)}}$ \\
    ODE solver = midpoint, 6 steps \\
    optimizer = LBFGS \\
    $\alpha = 0$, $\beta = 0$
} & 
\makecell[tl]{
    $N_{parallel} = 10$ \\
    $N_{steps}=500$, burn=$100$ \\
    $\eta(i) = 5e^{-2}$ \\
    $s(i) = 1e^{-2}$, $\lambda = 0.1$ (S-curve), $0.05$ (Two-moon) \\
    $\omega = 0.99$ \\
    $\delta(i) = 1e^{-3}$ \\
    ODE solver = midpoint, 6 steps \\
    $\alpha = 0$, $\beta = 0$
} & 
\makecell[tl]{
    $\lambda(\tau) = 1$ \\
    $b = \{1, 3, 10\}$ \\
    ODE solver = euler, 300 steps
} \\
\midrule
\textbf{KS} & 
\makecell[tl]{
    $N_{optim}= 10$ \\
    ODE solver = midpoint, 6 steps \\
    optimizer = LBFGS \\
    $\alpha = 0$, $\beta = 0$
} & 
\makecell[tl]{
    $N_{parallel} = 2$ \\
    $N_{steps}=300$, burn=$100$ \\
    $\eta(i) = 1e^{-2}$ \\
    $s(i) = 1e^{-3}$, $\lambda =  1e^{-3}$ \\
    $\omega = 0.99$ \\
    $\delta(i) = 1e^{-3}$ \\
    ODE solver = midpoint, 6 steps \\
    $\alpha = 0$, $\beta = 0$
} & 
\makecell[tl]{
    $\lambda(\tau) = 1$ \\
    $b = 1$ \\
    ODE solver = euler, 300 steps}\\
\midrule
\textbf{Turb} & 
\makecell[tl]{
    $N_{optim}= 100$ \\
    ODE solver = midpoint, 6 steps \\
    optimizer = LBFGS \\
    $\alpha = 0$, $\beta = 0$
} & 
\makecell[tl]{
    $N_{parallel} = 20$ \\
    $N_{steps}=600$, burn=$100$ \\
    $\eta(i) = 5e^{-2}$ \\
    $s(i) = 1e^{-3}$, $\lambda =  5e^{-6}$ \\
    $\omega = 0.99$ \\
    $\delta(i) = 1e^{-3}$ \\
    ODE solver = midpoint, 6 steps \\
    $\alpha = 0$, $\beta = 0$
    } &
\makecell[tl]{
    $\lambda(\tau) = 1$ \\
    $b = 1$ \\
    ODE solver = euler, 300 steps
} \\
\bottomrule
\end{tabular}
\end{table}

\section{Effect of the conditioning scale on Grad and Grad-Free methods}
\label{app:b-comp-study}
We examine the sensitivity of guided-transport conditioning to the guidance scale $b$ for the Grad and Grad-Free methods. We use the S-curve toy distribution with measurement operator $\mathcal{F}_1(\cdot)$ as a controlled setting where posterior geometry can be directly visualized.
\begin{figure}[!htp]
    \centering
    \includegraphics[width=0.8\linewidth]{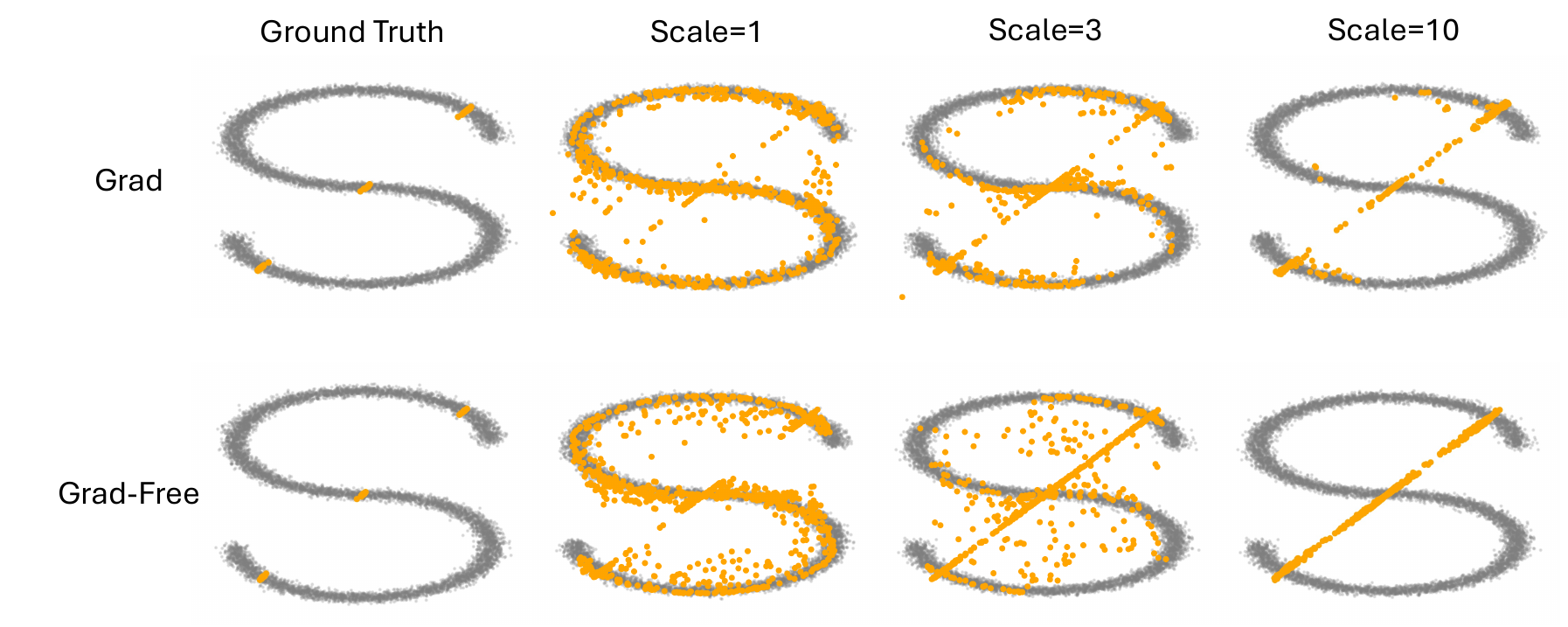}
    \caption{Effect of the guidance scale \(b\) on conditional samples for the S-curve toy problem with measurement operator $\mathcal{F}_1(\cdot)$ measurement function}
    \label{fig:scale_ablation}
\end{figure}
As shown in Figure~\ref{fig:scale_ablation}, the guidance scale $b$ controls a clear trade-off between measurement assimilation and prior fidelity. For small $b$, the conditional samples remain close to the learned manifold (gray prior support) but exhibit weak conditioning, with many samples behaving similarly to unconditional generation. As $b$ increases, the likelihood injection becomes stronger and measurement assimilation improves; however, this comes at the cost of prior fidelity. In particular, at moderate to large $b$, an increasing fraction of samples depart from the learned manifold, indicating that the guided dynamics can be driven into regions poorly supported by the pretrained prior. This effect is most pronounced for Grad-Free, and Grad exhibits the same trend but with fewer extreme outliers. Overall, the ablation shows that guidance-based conditioning is not robust, and its performance depends critically on tuning $b$, and large $b$ can improve measurement fit while degrading prior plausibility, which is problematic in scientific inverse problems.

\clearpage
\section{Reconstruction from sparse sensors for KS equation}
\label{sec:KS-sparse-sensor}

\begin{figure}[H]
    \centering
    \includegraphics[width=0.9\linewidth]{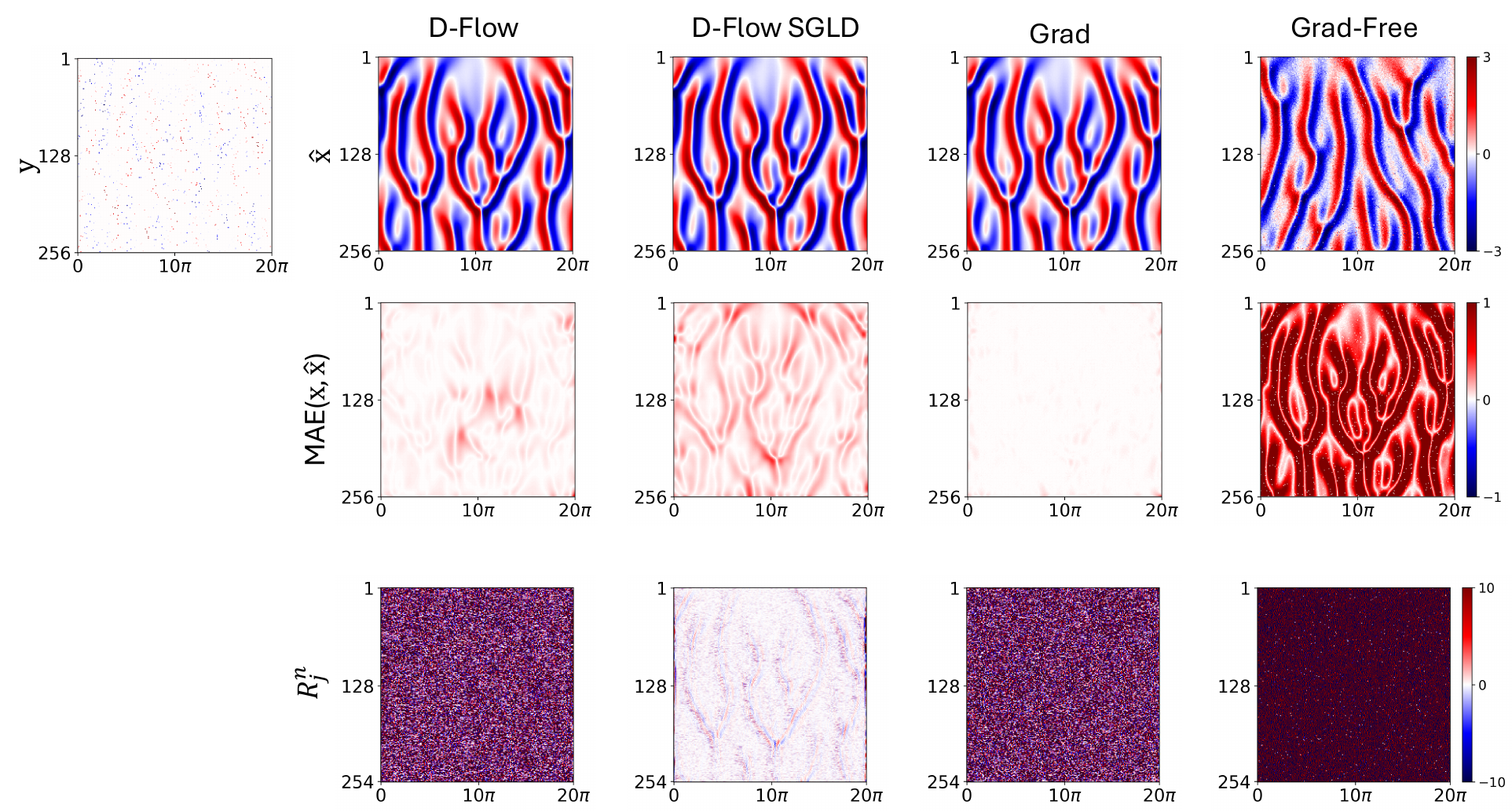}
    \caption{KS sparse sensor reconstruction with noise-free measurements. \textbf{(top row)} The ground truth measurement is shown in the first column, followed by a conditionally generated sample from each method. The region below the white line ($t>r$) is the forecasting portion. \textbf{(middle row)} Pointwise data assimilation error (MAE). \textbf{(bottom row)} The PDE residual, $R_j^n$, computed over the entire generated field to assess physical plausibility.}
    \label{fig:ks_meas_sparse_noise_free}
\end{figure}

\begin{figure}[H]
    \centering
    \includegraphics[width=0.9\linewidth]{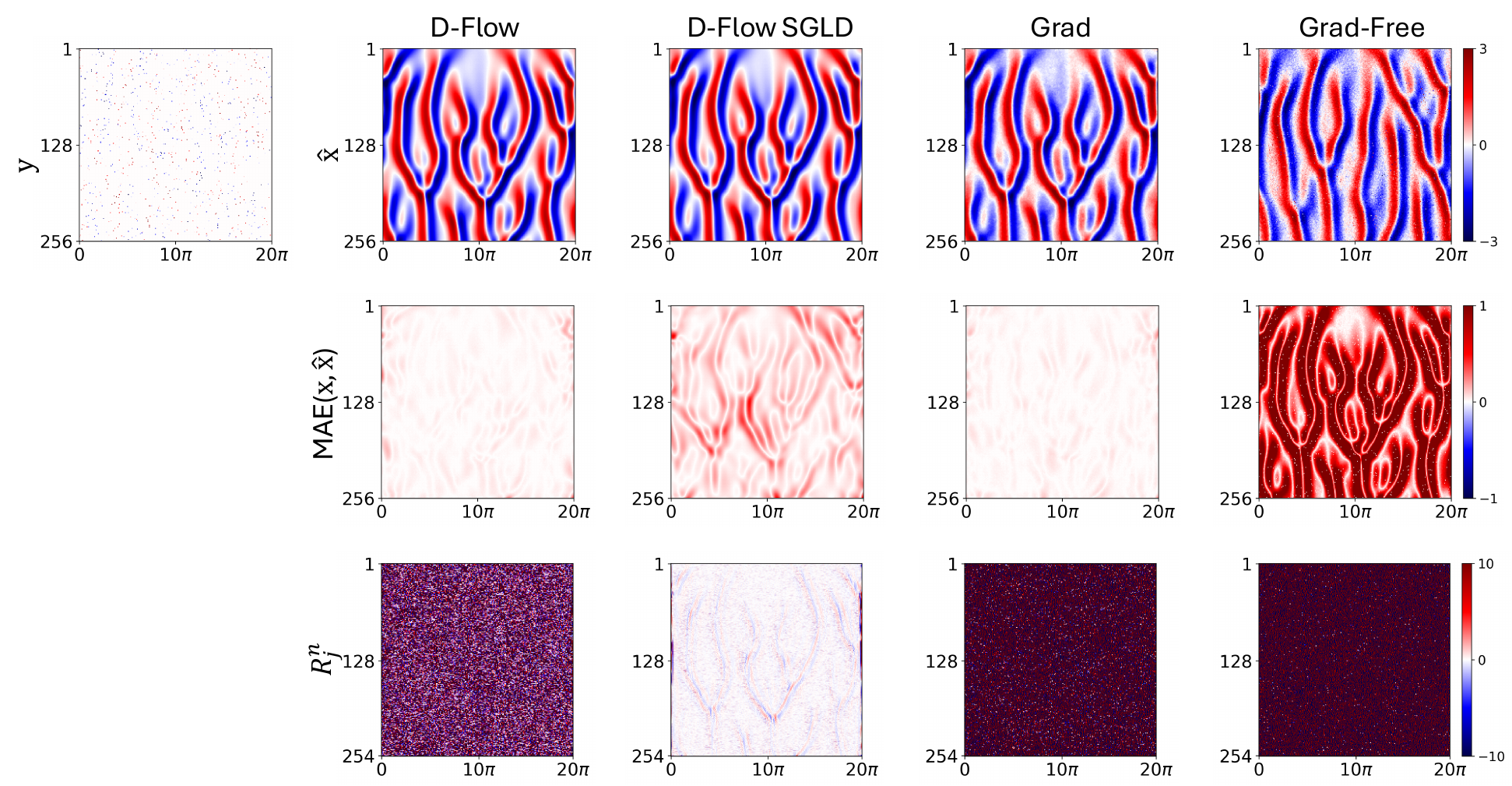}
    \caption{KS sparse sensor reconstruction with noisy measurements ($10\%$). \textbf{(top row)} The ground truth measurement is shown in the first column, followed by a conditionally generated sample from each method. The region below the white line ($t>r$) is the forecasting portion. \textbf{(middle row)} Pointwise data assimilation error (MAE). \textbf{(bottom row)} The PDE residual, $R_j^n$, computed over the entire generated field to assess physical plausibility.}
    \label{fig:ks_meas_sparse_noise}
\end{figure}

For the sparse-sensor setting, we condition on 1000 pointwise measurements sampled uniformly at random from the space--time domain. Given a pretrained prior that captures the KS solution manifold, these sparse observations can strongly constrain the conditional distribution around the underlying trajectory. Figures~\ref{fig:ks_meas_sparse_noise_free} and \ref{fig:ks_meas_sparse_noise} show that all guidance methods except the Grad-free can recover visually plausible reconstructions consistent with the sensor data.

A closer comparison reveals important differences in assimilation and physics fidelity. D-Flow SGLD produces accurate reconstructions in both the noise-free and noisy cases and achieves the lowest PDE residual, indicating the best preservation of KS consistency. D-Flow also yields visually accurate reconstructions, but with higher residuals in some instances, consistent with sensitivity of deterministic source optimization to the conditioning instance and optimization landscape. Grad attains low MAE, indicating strong data assimilation, but its residual increases substantially under noisy measurements, suggesting overfitting to corrupted observations at the expense of the governing PDE. Grad-Free performs worst: it exhibits high MAE even in the noise-free case and deteriorates further with noise, producing visibly corrupted reconstructions and very large residuals, highlighting limited robustness in this sparse-sensor regime.


\section{Additional results for wall-bounded turbulence case}
\label{sec:additonal-results}

\begin{figure}[htp!]
    \centering
    \includegraphics[width=0.9\linewidth]{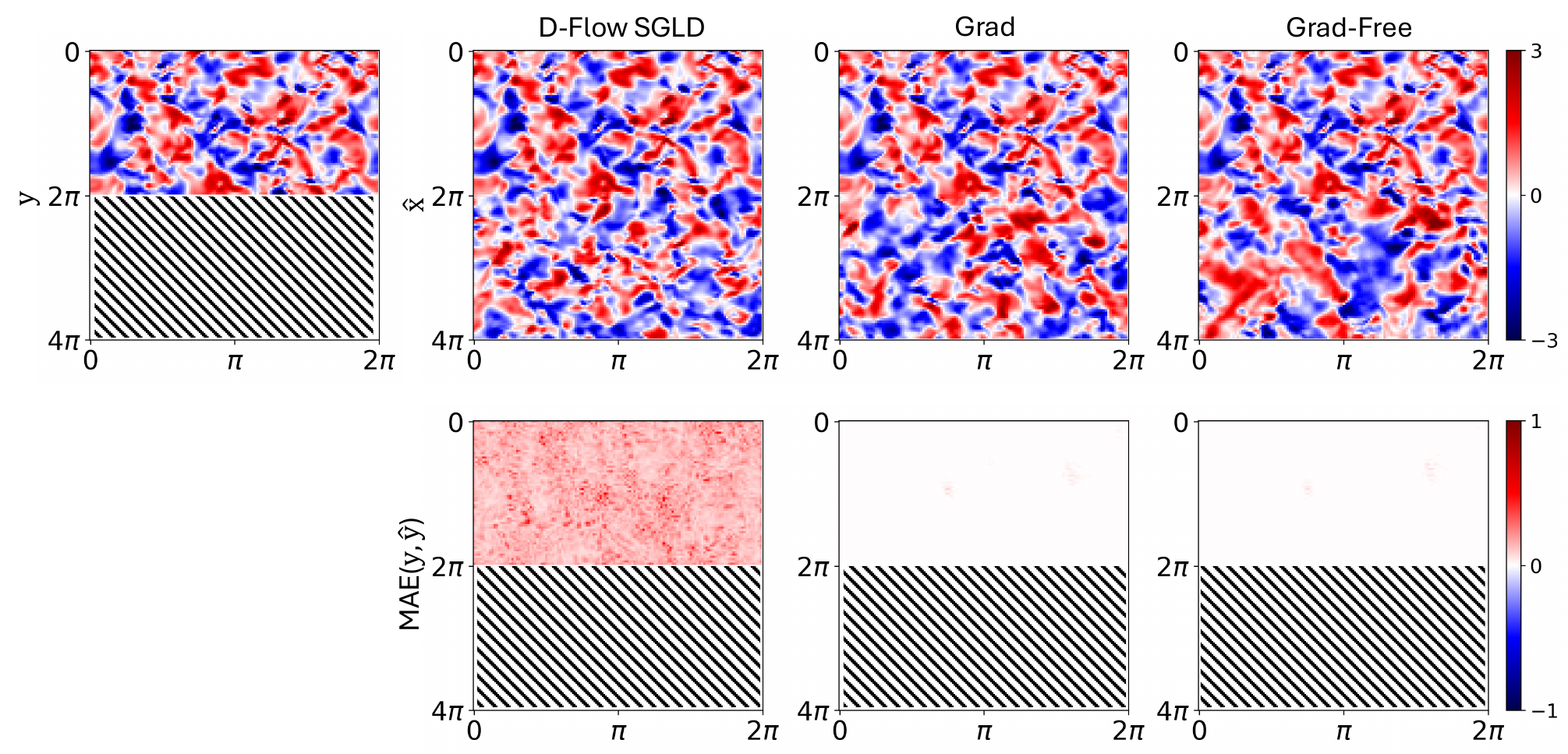}
    \caption{Turbulence inpainting ($u_2$) with noise-free measurements. Top: the observed field (first column) and one representative conditional sample from each method; the hatched region ($x>2\pi$) indicates the missing/inpainted portion. Bottom: pointwise MAE on the observed region.}
    \label{fig:turb_noise_free_contour_v}
\end{figure}

\begin{figure}[H]
    \centering
    \includegraphics[width=0.9\linewidth]{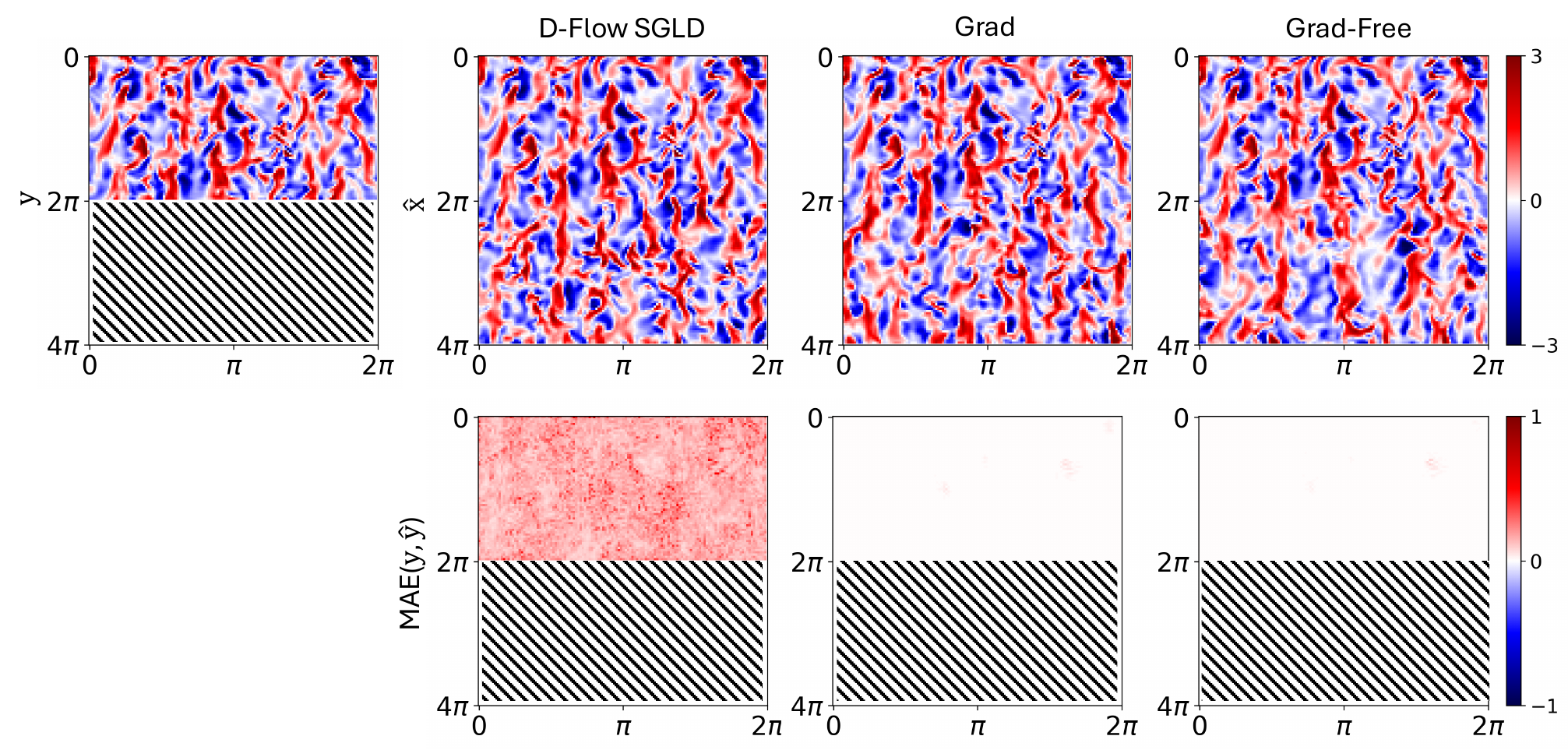}
    \caption{Turbulence inpainting ($u_3$) with noise-free measurements. Top: the observed field (first column) and one representative conditional sample from each method; the hatched region ($x>2\pi$) indicates the missing/inpainted portion. Bottom: pointwise MAE on the observed region.}
    \label{fig:turb_noise_free_contour_w}
\end{figure}

\begin{figure}[H]
    \centering
    \includegraphics[width=0.9\linewidth]{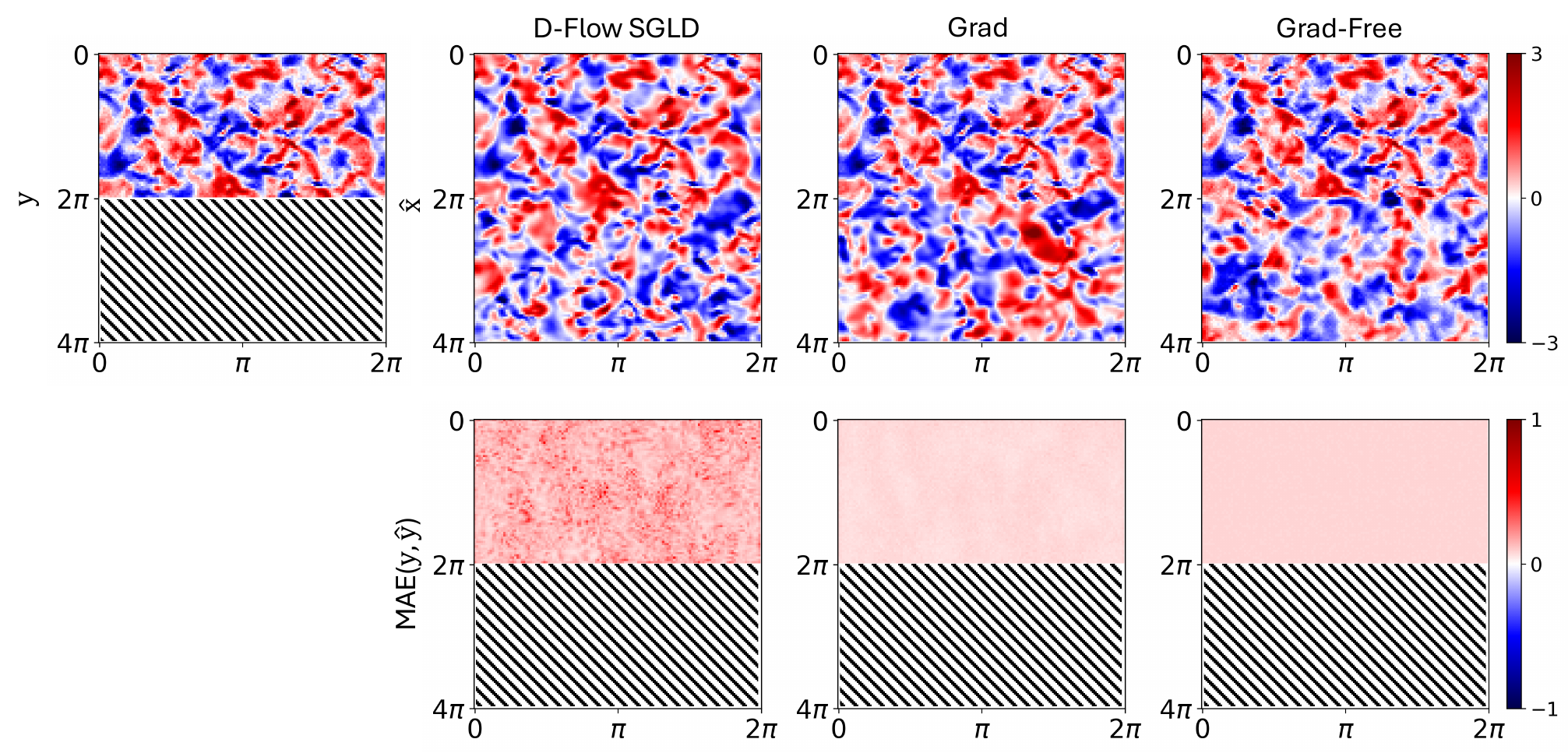}
    \caption{Turbulence inpainting ($u_2$) with noisy measurements ($10\%$). Top: the observed field (first column) and one representative conditional sample from each method; the hatched region ($x>2\pi$) indicates the missing/inpainted portion. Bottom: pointwise MAE on the observed region.}
    \label{fig:turb_noise_contour_v}
\end{figure}

\begin{figure}[H]
    \centering
    \includegraphics[width=0.9\linewidth]{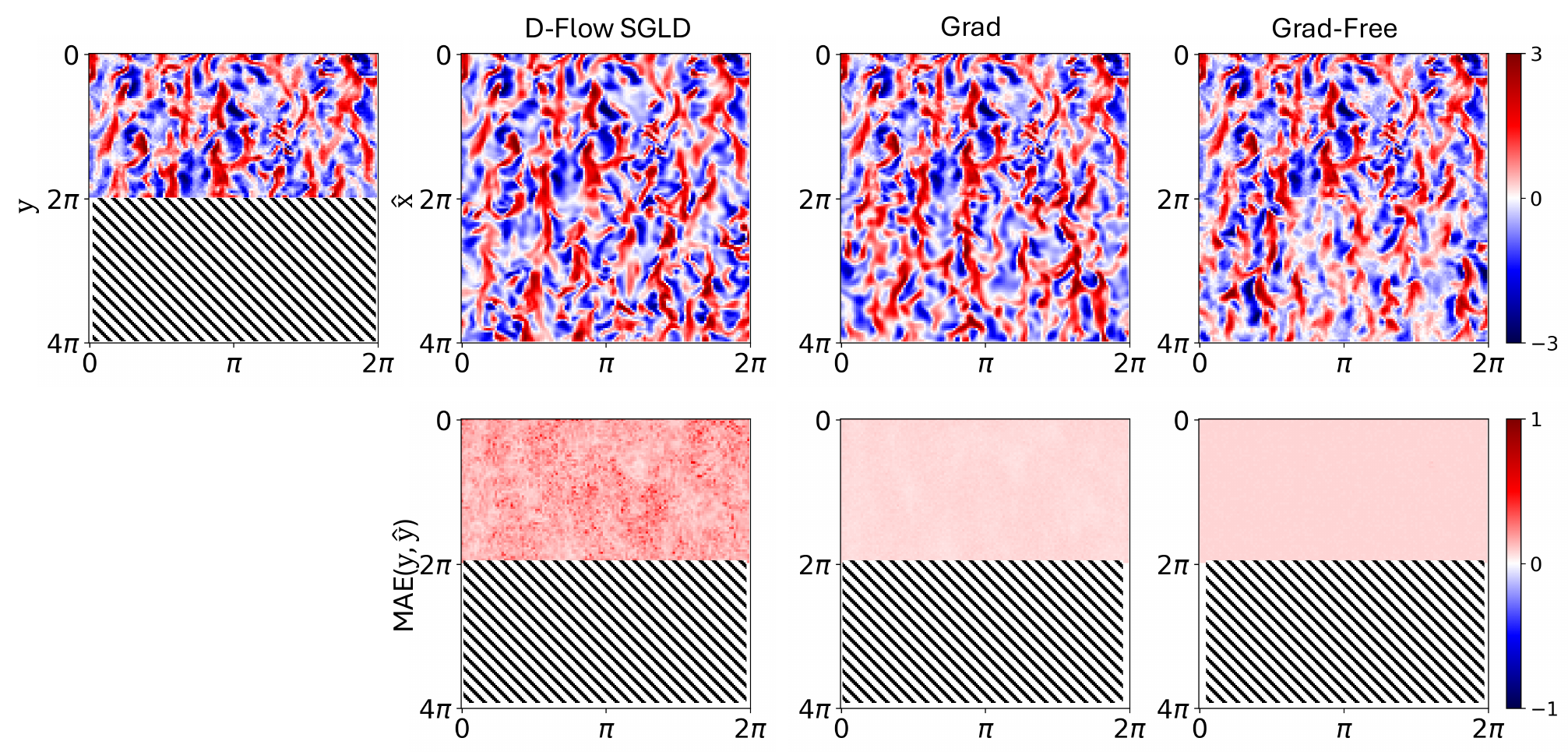}
    \caption{Turbulence inpainting ($u_3$) with noisy measurements ($10\%$). Top: the observed field (first column) and one representative conditional sample from each method; the hatched region ($x>2\pi$) indicates the missing/inpainted portion. Bottom: pointwise MAE on the observed region.}
    \label{fig:turb_noise_contour_w}
\end{figure}

\begin{figure}[H]
\centering
\subfloat[Noise-free case]{\includegraphics[width=0.9\textwidth]{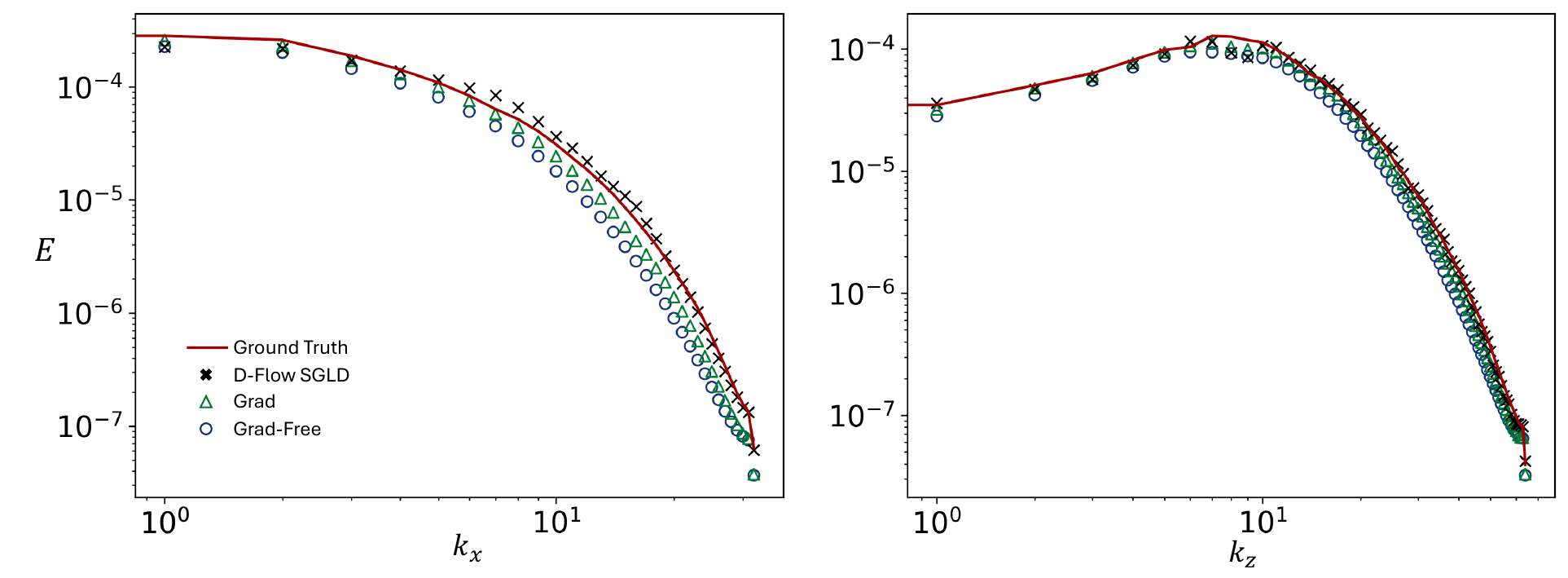}}\\
\subfloat[Noise corrupted case] {\includegraphics[width=0.9\textwidth]{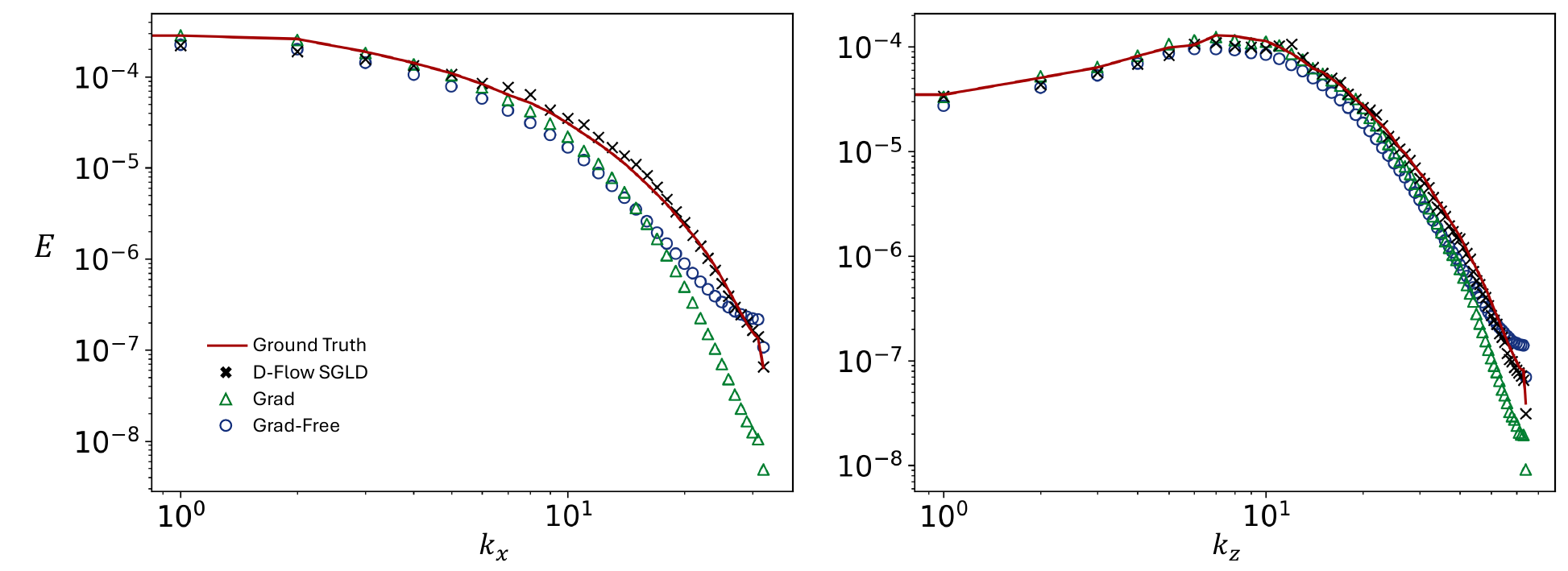}}\\
\caption{Comparison of the streamwise (right) and spanwise (left) one-dimensional energy spectrum for the wall-normal velocity component $u_2$ at wall-normal location of $y^+ =40$.}
\label{fig:turb_energy_spectrum_v}
\end{figure}

\begin{figure}[H]
\centering
\subfloat[Noise-free case]{\includegraphics[width=0.9\textwidth]{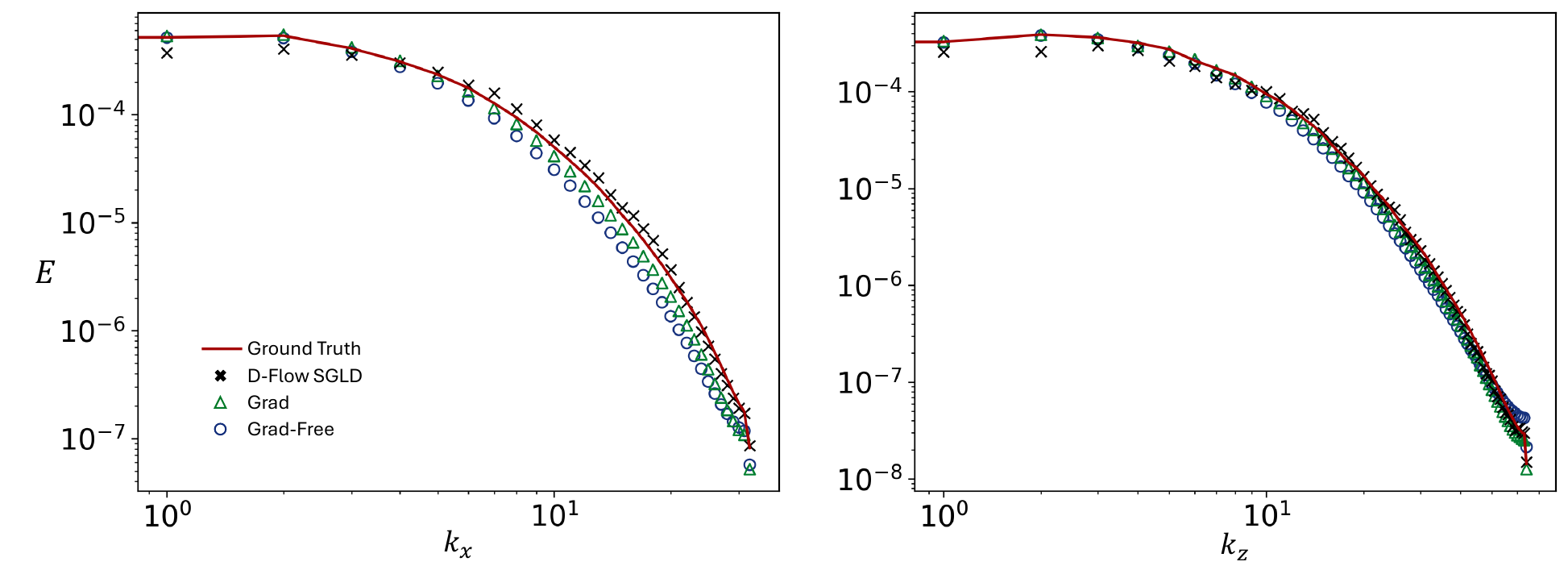}}\\
\subfloat[Noise corrupted case] {\includegraphics[width=0.9\textwidth]{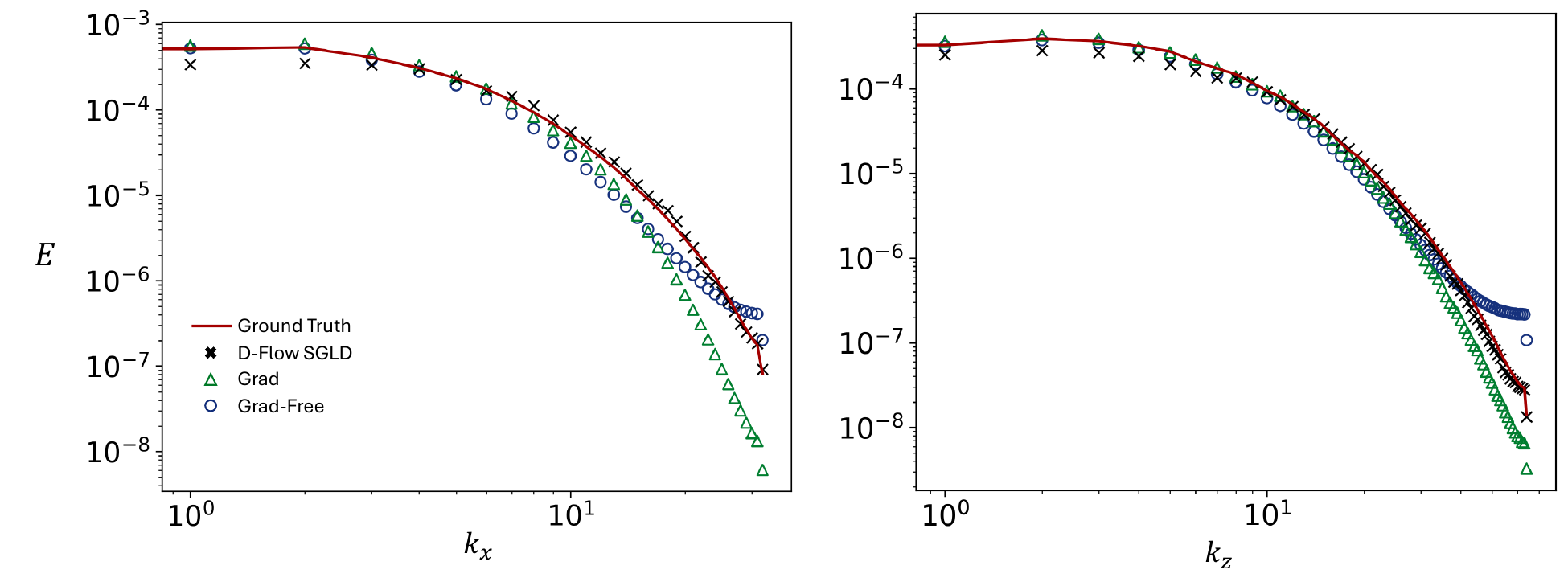}}\\
\caption{Comparison of the streamwise (right) and spanwise (left) one-dimensional energy spectrum for the spanwise velocity component $u_3$ at wall-normal location of $y^+ =40$.}
\label{fig:turb_energy_spectrum_w}
\end{figure}

For completeness, the results for the wall-normal ($u_2$) and spanwise ($u_3$) velocity components are presented in Figures~\ref{fig:turb_noise_free_contour_v} through \ref{fig:turb_energy_spectrum_w}, which show identical trends observed for $u_1$. 

\bibliographystyle{elsarticle-num}
\bibliography{ref}

\end{document}